\documentclass[10pt,twocolumn,letterpaper]{article}

\usepackage[pagenumbers]{cvpr} 

\usepackage{amsfonts}
\usepackage{graphicx}
\usepackage{pifont}
\usepackage{tabularx}
\usepackage{multicol}
\usepackage{multirow}
\usepackage{threeparttable}
\usepackage{caption}
\usepackage{xcolor}
\usepackage{makecell}
\usepackage{tcolorbox}
\usepackage{booktabs}
\usepackage{bm}

\definecolor{cvprblue}{rgb}{0.21,0.49,0.74}
\usepackage[pagebackref,breaklinks,colorlinks,allcolors=cvprblue]{hyperref}

\title{Concept Corrector: Erase concepts on the fly for text-to-image diffusion models}

\author{Zheling Meng$^{1,2,}$,\;
Bo Peng$^{1,}$,\;
Xiaochuan Jin$^{1,2}$,\;
Yueming Lyu$^{3}$,\;
Wei Wang$^{1}$,\; 
Jing Dong$^{1}$,\; 
Tieniu Tan$^{1,3}$\;\\
\normalsize{$^{1}$ NLPR, Institute of Automation CAS, \quad $^{2}$ School of Artificial Intelligence UCAS, \quad $^{3}$ Nanjing University,}\\
{\tt\small zheling.meng@cripac.ia.ac.cn, jdong@nlpr.ia.ac.cn}
}

\begin{document}

\twocolumn[{
    \renewcommand\twocolumn[1][]{#1}
    \maketitle
    \vspace{-1cm}
    
\begin{center}
    \captionsetup{type=figure}
    \includegraphics[scale=0.35, trim=20 0 0 510,clip]{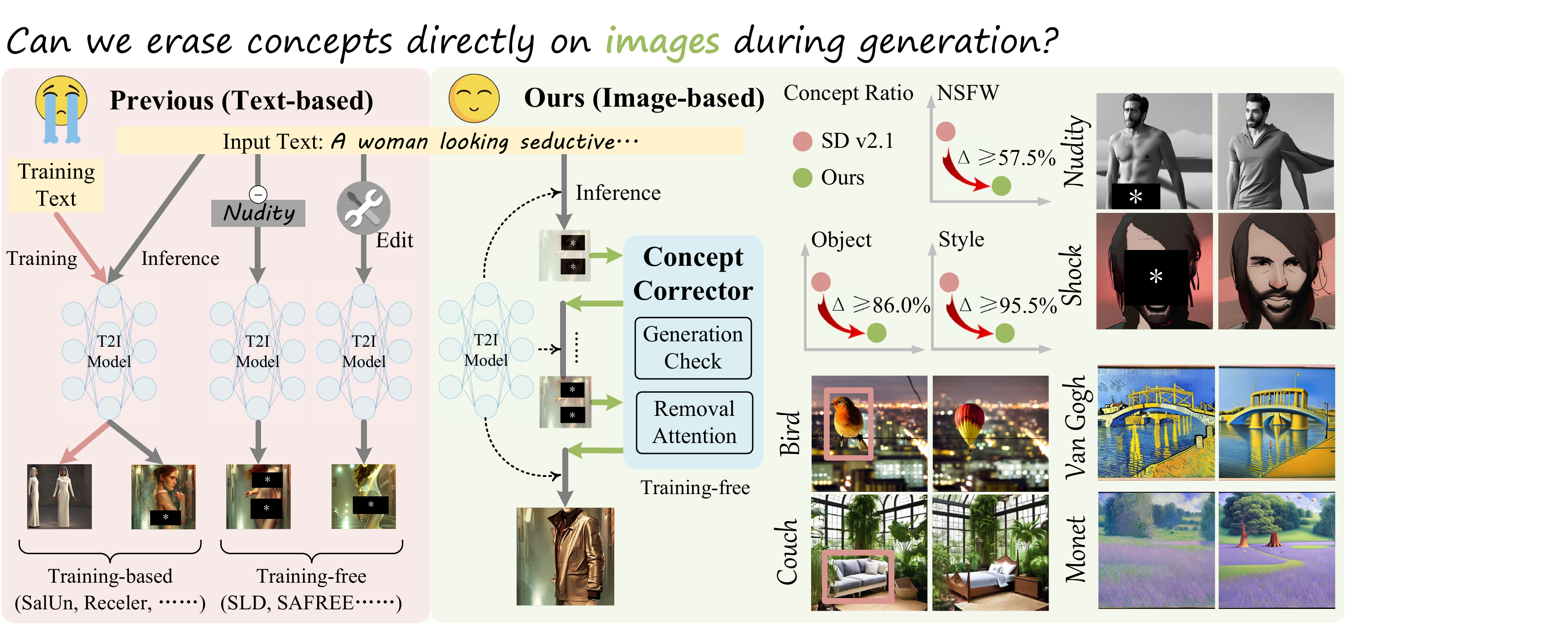}
    \captionof{figure}{We present \textbf{Concept Corrector}, a training-free method to erase concepts based on intermediate-generated images during the text-to-image diffusion process. It integrates the \textbf{Generation Check Mechanism} alongside \textbf{Concept Removal Attention}, to first check target concepts during generation and then erase them in the subsequent generations. Please refer to Appendix J for more visualizations.}
    \label{fig: example_main}
\end{center}
}]


\begin{abstract}

Text-to-image diffusion models have demonstrated the underlying risk of generating various unwanted content, such as sexual elements. To address this issue, the task of concept erasure has been introduced, aiming to erase any undesired concepts that the models can generate. Previous methods, whether training-based or training-free, have primarily focused on the input side, i.e., texts. However, they often suffer from incomplete erasure due to limitations in the generalization from limited prompts to diverse image content. In this paper, motivated by the notion that concept erasure on the output side, i.e., generated images, may be more direct and effective, we propose \textbf{Concept Corrector}. It checks target concepts based on visual features provided by final generated images predicted at certain time steps. Further, it incorporates Concept Removal Attention to erase generated concept features. It overcomes the limitations of existing methods, which are either unable to remove the concept features that have been generated in images or rely on the assumption that the related concept words are contained in input prompts. In the whole pipeline, our method changes no model parameters and only requires a given target concept as well as the corresponding replacement content, which is easy to implement. To the best of our knowledge, this is the first erasure method based on intermediate-generated images, achieving the ability to erase concepts on the fly. The experiments on various concepts demonstrate its impressive erasure performance. \href{https://github.com/RichardSunnyMeng/ConceptCorrector}{Code}.

\end{abstract}    
\section{Introduction}
\label{sec: introduction}

In recent years, text-to-image diffusion models \cite{gal2022image, mou2024t2i, nichol2021glide, rombach2022high,  ruiz2023dreambooth, saharia2022photorealistic}, such as Stable Diffusion \cite{rombach2022high}, have developed rapidly, attracting widespread attention from academia and industry. They are usually pre-trained on large-scale datasets and then fine-tuned on downstream data as needed. Benefiting from the diversity of datasets, text-to-image diffusion models can acquire rich visual features of entities in the physical world and associate them with the text modality. They can not only generate high-fidelity images but also possess excellent text conditioning ability, thereby producing image content that aligns with user intentions \cite{zhang2023text, cao2024controllable}. However, everything has two sides. Large-scale training datasets inevitably contain undesired content, such as nudity, blood, copyright, etc. It endangers the models themselves with the capability to generate such content \cite{qu2023unsafe}, threatening individual rights and social harmony. 

The task of \emph{concept erasure} has been proposed to address this issue \cite{schramowski2023safe, kumari2023ablating, gandikota2023erasing}. Its goal is to prevent models from generating content related to certain concepts. Intuitively speaking, if the generation path from the textual inputs to the target visual concepts can be cut off, the resultant images, being guided by those texts, would naturally exclude the presence of these concepts. Researchers have explored various methods based on this idea, which can be categorized into training-based and training-free methods. Within the area of training-based methods, CA \cite{kumari2023ablating} and ESD \cite{gandikota2023erasing} fine-tune the generation distributions conditioned on the texts containing target concepts. Instead of gathering texts, Bui et al. \cite{, bui2024removing} and Meng et al. \cite{meng2024dark} incorporate a learnable prompt in the training and other works \cite{pham2024robust, kim2024race, zhang2024defensive, huang2023receler} introduce adversarial training to improve the robustness of concept erasure. UCE \cite{gandikota2024unified}, MACE \cite{lu2024mace}, and RECE \cite{gong2024reliable} edit cross-attention weights by aligning the keys and values of target concepts to others. There are also some methods \cite{fan2023salun, yang2024pruning, chavhan2024conceptprune} ablating model parameters based on their sensitivity to related prompts. While these methods fine-tune diffusion models, Latent Guard \cite{liu2025latent} and GuardT2I \cite{yang2024guardt2i} propose to fine-tune the text encoders. Within the area of training-free methods, Negative Guidance \cite{rombach2022high} and SLD \cite{schramowski2023safe} use concept texts to steer generation in the opposite direction. SAFREE \cite{yoon2024safree} maps the prompt embeddings to those of target concepts and removes the corresponding components.

While most studies explore text-based concept erasure methods, little attention has been paid to accomplishing this task based on the output of generative models, i.e. generated images. Text-based erasure methods often face the challenge of \emph{prompt generalization} \cite{tsai2023ring, zhang2023generate, yang2024mma, chin2023prompting4debugging}. It arises due to the difficulty in using prompts, which are collected or learned by these methods, to comprehensively cover the diverse image content associated with target concepts. Consequently, the effectiveness of the erasure is limited. Consider a simple example. The prompt ``A woman looking seductive'' does not explicitly convey the meaning of nudity. However, as demonstrated in Fig.\ref{fig: example_main}, it can prompt the erased models to generate an image of a naked woman. If the focus is shifted to images, the target of erasure becomes more directly addressed, potentially leading to a more effective erasure performance. A common practice is to incorporate a safety checker following the generation process to filter out unwanted content like in Stable Diffusion \cite{rombach2022high}, which works well but cannot correct generated content to obtain images like text-based erasure methods. 

In this paper, we aim to actively intervene intermediate images in the generation to reliably erase target concepts from final images. Using the diffusion generation theory, we note that final images predicted at certain time steps present enough structure and detail features for a detector to check concepts. Once concepts are detected, we consider how to erase them. Existing solutions like prompt editing \cite{hertzprompt} and negative guidance \cite{ho2021classifier, schramowski2023safe} cannot remove the concept features from images, causing target concepts to still exist. Some methods like Receler \cite{huang2023receler} suppress concept features within the attention layers. However, they rely on concept words in input prompts, which are difficult to anticipate in advance or accurately capture during inference due to the diversity and implicitness of languages. Thus, we propose Concept Removal Attention, a variation of the cross attention mechanism. It erases generated concept features by giving the names of target concepts and negative concepts and perturbing generated features. These efforts form our method \textbf{Concept Corrector}. 

Compared with previous methods, our method has significant advantages in terms of \textbf{reliability}. Firstly, it checks concepts in images, thereby providing a more direct assessment of concepts. Secondly, it does not require input prompts to contain explicit concept words, solving the problem of prompt generalization. Moreover, all parameters remain unchanged, which protects model knowledge. In the experiments, we evaluate the erasure performance using user prompts and adversarial prompts. The evaluated concepts include Not-Safe-For-Work, objects, and painting styles. As shown in Fig.\ref{fig: example_main}, Tab.\ref{tab: main results} and Tab.\ref{tab: adversarial results}, our method achieves impressive erasure performance, significantly reducing the generation of most concepts to within 5\% while other methods are still far from this level of performance. The contributions of this paper can be outlined as follows.

\begin{itemize}
    \item We carefully analyze the feasibility of using intermediate images in the generation for checking target concepts. 
    \item We propose Concept Removal Attention to erase concept features in the generation. It changes no model parameters and only requires the names of target concepts and negative concepts, making erasure easy to implement.
    \item The above explorations forms Concept Corrector, a straightforward but effective method to erase concepts on the fly. To our knowledge, this is the first work to achieve erasure based on intermediate-generated images.
    \item The experiments and visualizations demonstrate the impressive effectiveness of our method. A series of ablation experiments are conducted to discuss each component.
\end{itemize}

\section{Related Work}
\label{sec: related work}

\subsection{Training-based Erasure}

We first summary the training-based concept erasure methods. Here, the word "training-based" generally refers to methods that change model parameters in various ways.

\textbf{Generative distribution alignment.} Concept Ablating (CA) \cite{kumari2023ablating} matches the generative distribution of a target concept to the distribution of an anchor concept. Erasing Stable Diffusion (ESD) \cite{gandikota2023erasing} fine-tunes the distribution of a target concept to mimic the negatively guided ones. While CA and ESD align the predicted noises, Forget-Me-Not (FMN) \cite{zhang2023forget} suppresses the activation of concept-related content in the attention layers. Considering the gap between the visual and textual features in text-to-image diffusion models, Knowledge Transfer and Removal \cite{bui2024removing} is proposed to replace collected texts with learnable prompts. Dark Miner \cite{meng2024dark} also conveys this idea. Adversarial training is also introduced for robustness erasure \cite{pham2024robust, kim2024race, zhang2024defensive, huang2023receler}.

\textbf{Parameter editing.} Unified Concept Editing (UCE) \cite{gandikota2024unified} formalizes the erasure task by aligning the projection vectors of target concepts to those of anchor concepts in the attention layers. It derives a closed-form solution for the attention parameters under this objection and edits the model parameters directly. Based on UCE, Reliable and Efficient Concept Erasure (RECE) \cite{gong2024reliable} introduces an iterative editing paradigm for a more thorough erasure. Mass Concept Erasure  (MACE) \cite{lu2024mace} leverages the closed-form parameter editing along with parallel LoRAs \cite{hu2021lora} to enable multiple concept erasure.

\textbf{Model pruning.} Previous studies find that certain concepts activate specific neurons in a neural network \cite{wang2022finding}. Yang et al. \cite{yang2024pruning} selectively prune critical parameters related to concepts and empirically confirm
its superior performance. SalUn \cite{fan2023salun} proposes a new metric named weight saliency and utilizes the gradient of a forgetting loss to ablate the salient parameters. Relying on the forward process, ConceptPrune \cite{chavhan2024conceptprune} identifies activated neurons of the feed-forward layers and zeros them out.

\textbf{Text encoder fine-tuning.} The methods mentioned above modify the parameters of diffusion models, ignoring the text encoder, another important component in the generation process. Latent Guard \cite{liu2025latent} learns an embedding mapping layer on top of the text encoder to check the presence of concepts in the prompt embeddings. GuardT2I \cite{yang2024guardt2i} fine-tunes a Large Language Model to convert prompt embeddings into natural languages and analyze their intention, which helps determine the presence of concepts in generated images under the guidance of these prompts.

\subsection{Training-free Erasure}

The training-free methods focus on using the inherent ability of diffusion models to prevent the generation of concept-related content. Safe Latent Diffusion (SLD) \cite{schramowski2023safe} is a pioneering work in this field. SLD proposes safety guidance. It extends the generative diffusion process by subtracting the noise conditioned on target concepts from the noise predicted at each time step. Recently, SAFREE \cite{yoon2024safree} constructs a text embedding subspace using target concepts and removes the components of input embeddings in the corresponding subspace. Further, SAFREE fuses the latent images conditioned on the initial and processed embeddings in the frequency domain.

\begin{figure*}[]
    \centering
    \includegraphics[width=1\linewidth, trim=0 0 0 40,clip]{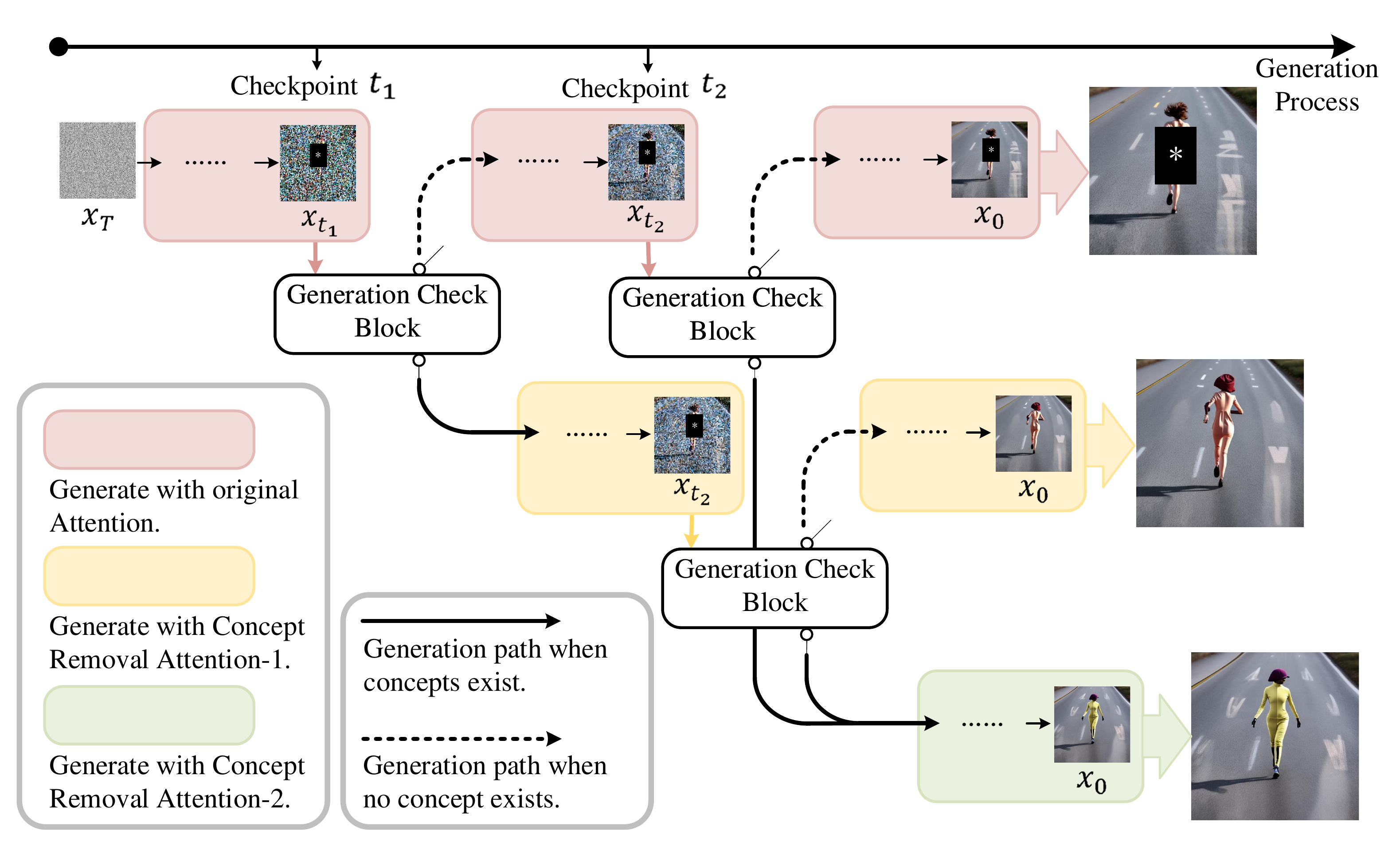}
    \caption{The generation pipeline of text-to-image diffusion models with our proposed Concept Corrector.}
    \label{fig: pipeline}
\end{figure*}

\section{Preliminaries}
\subsection{Latent Diffusion Models}
\label{sec: ldm}

Diffusion models \cite{sohl2015deep, ho2020denoising} iteratively estimate and remove the noise from the sampled Gaussian noise, yielding images after $T$ steps. They act as noise predictors conditioned on the time step $t$. Latent diffusion models \cite{rombach2022high} execute the process in a latent space and then decode generated latent images into pixel space. Text-to-image diffusion models \cite{ho2021classifier} incorporate a prompt $p$ as a condition for the noise prediction, achieving the generation of images aligned with $p$.

Let $\epsilon_\theta(z_t, t, p)$ denote a noise predictor with the parameters $\theta$. At the time step $t (T \geq t > 0)$, the estimated noise:
\begin{equation}
    \hat{\epsilon}_\theta(z_t, t, p) = \epsilon_\theta(z_t, t, \varnothing) + \gamma (\epsilon_\theta(z_t, t, p) - \epsilon_\theta(z_t, t, \varnothing)),
\end{equation}
where $\gamma$ is a guidance scale and $\varnothing $ denotes an empty prompt. When $t=T$, $z_t$ denotes a sampled Gaussian noise. Then $z_{t-1}$ follows a Gaussian distribution $N(z_{t-1}|\bm{\mu}_{t-1}, \sigma_{t-1}^2\mathbf{I})$:
\begin{equation}
    p(z_{t-1}|z_t, p) = N(z_{t-1};\frac{1}{\sqrt{\alpha_t}}(z_t-\frac{\beta_t}{\overline{\beta}_t}\hat{\epsilon}_\theta(z_t, t, p)), \frac{\overline{\beta}_{t-1}^2\beta_t}{\overline{\beta}_t^2}\mathbf{I}),
\end{equation}
where $\beta_t$ is a scheduled noise variance, $\alpha_t={1-\beta_t}$, $\overline{\alpha}_t=\alpha_1...\alpha_t$, and $\overline{\beta}_t=\sqrt{1-\overline{\alpha}_t}$. According to the Markov theory, we can derive $\hat{z}_0$ to predict the final denoising result $z_0$ \cite{sohl2015deep}:
\begin{equation}
\label{eq: predict z0}
    \hat{z}_0 = \frac{1}{\sqrt{\overline{\alpha}}_t}(z_t-\overline{\beta}_t\hat{\epsilon}_\theta(z_t, t, p)).
\end{equation}
We call Eq.\ref{eq: predict z0} the predictability of diffusion models. The intermediate image $x_t$ and the predicted final image $\hat{x}_0$ are decoded from $z_t$ and $\hat{z}_0$ by the latent decoder respectively.

\subsection{Cross-Attention Mechanism}
\label{sec: attention}
Cross-attention is a key mechanism to achieve text-conditioning image generation. Usually, there are multiple parallel heads in the attention layers. For each attention head, the attention function $Attn(z_t, p)$ is defined as:
\begin{equation}
\label{eq: attention}
    Attn(z_t, p) = Softmax(\frac{QK^T}{\sqrt{d}})V,
\end{equation}
where the query $Q=W_qz_t$, the key $K=W_k\tau_\theta(p)$, the value $V=W_v\tau_\theta(p)$, $W_q$, $W_k$, and $W_v$ are the projection  matrix, and $\tau_\theta(p)$ denotes the text encoder. $d$ denotes the dimension of the features, $Q\in \mathbb{R}^{M\times d}$, $K\in \mathbb{R}^{N\times d}$, $V\in \mathbb{R}^{N\times d}$, and $Softmax(\frac{QK^T}{\sqrt{d}})\in\mathbb{R}^{M\times N}$ where $M$ is the pixel length of $z_t$ and $N$ is the token length of $p$.

\section{Methods}
\label{sec: methods}

\subsection{Overview}
Fig.\ref{fig: pipeline} illustrates the generation pipeline of text-to-image diffusion models with our proposed Concept Corrector. This pipeline introduces two checkpoints, $t_1$ and $t_2$, within the diffusion process. As the time step $t$ reaches the checkpoints, the intermediate result $x_t$ is directed into the \textbf{Generation Check Block}. It checks whether any content related to target concepts is present. If such content is absent, the generation proceeds uninterrupted in its original course. Conversely, if one or more concepts are detected, the subsequent generation integrates the proposed \textbf{Concept Removal Attention} as the cross-attention mechanism. Depending on which checkpoint, $t_1$ or $t_2$, triggers the check, Concept Removal Attention-1 or -2 is employed, respectively. They differ subtly in their approach to fusing attention features, thereby enhancing their adaptability to the generation preferences of different generation stages.

\subsection{Generation Check Mechanism}
\label{sec: generation check mechanism}

As mentioned in Sec.\ref{sec: ldm}, the final generation results can be predicted at intermediate time steps. We highlight that they provide rich visual features to check concepts during the generation. Some previous studies \cite{choi2022perception, jung2024latent} explore the generative traits of different diffusion stages. A common observation is that diffusion models initially generate global structures and then refine these with local details as the diffusion progresses. It inspires us to set two checkpoints to check concepts, leveraging the generated structures and details at different stages.

\begin{figure}[t]
    \centering
    \includegraphics[width=1.0\linewidth]{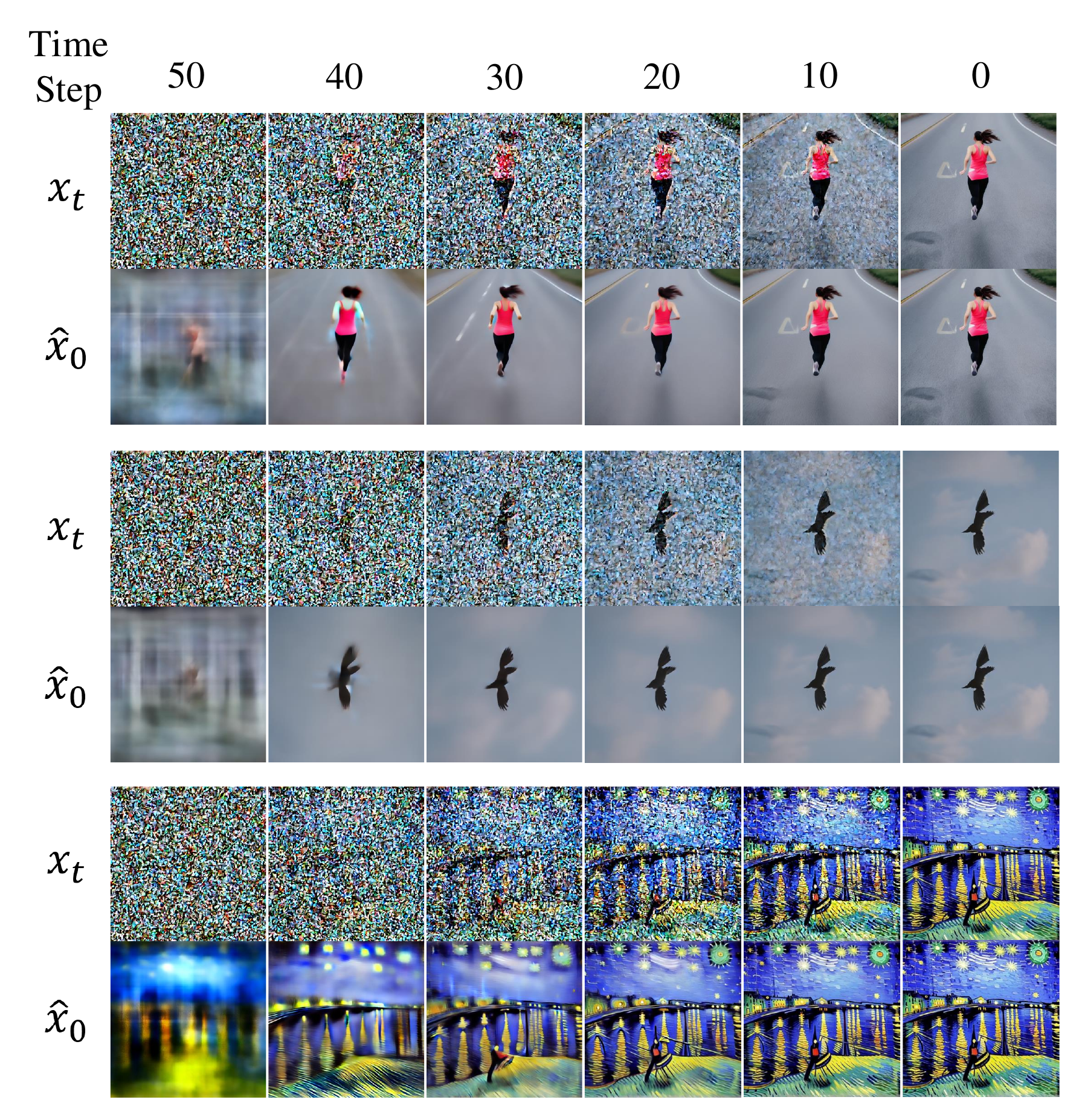}
    \caption{The examples for the intermediate results $x_t$ and the predicted final results $\hat{x}_0$ generated by Stable Diffusion v2.1. DDIM \cite{songdenoising} is the scheduler with 50 sampling steps. The top: bodies. The middle: birds. The bottom: Van Gogh's painting style.}
    \label{fig: mid results}
\end{figure}

To further elaborate on the above motivation, we provide some examples in Fig.\ref{fig: mid results}. In these illustrations, the intermediate images $x_t$ have heavy noise, rendering it challenging for humans to discern the content. On the contrary, the clarity of the predicted final images $\hat{x}_0$ undergoes a significant enhancement. Moreover, during the initial time steps, discernible structures such as bodies and birds become evident in $\hat{x}_0$. In subsequent time steps, additional details are progressively generated, aiding in the identification of more concepts, such as Van Gogh's painting style.

At each checkpoint, we set a Generation Check Block respectively. It receives the predicted final images $\hat{x}_0$ as input and uses a detector to decide whether they contain any target concepts. Please see Sec.\ref{sec: setting} for the implementation.


\subsection{Concept Removal Attention}

\begin{figure}[t]
    \centering
    \includegraphics[width=0.9\linewidth, trim=20 20 20 20,clip]{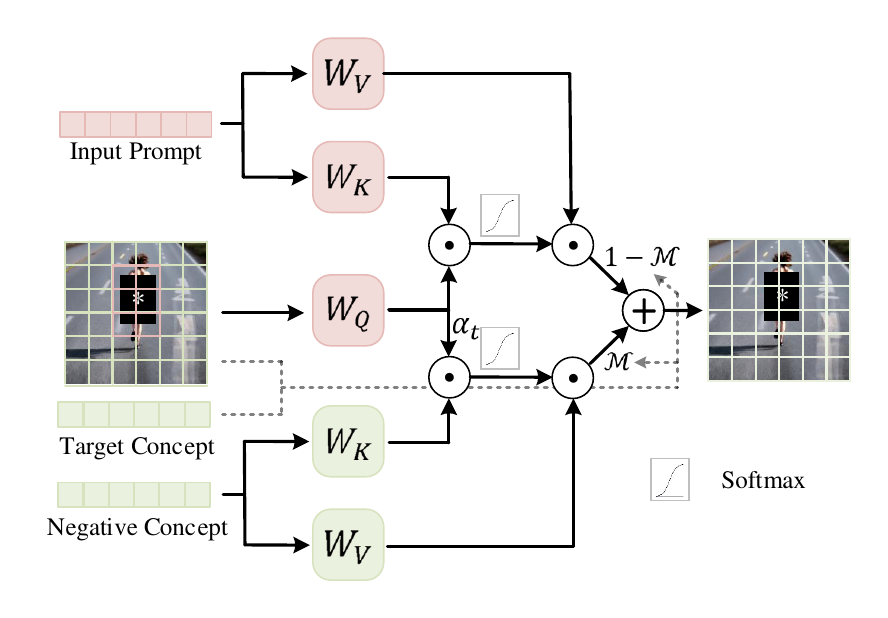}
    \caption{The framework of Concept Removal Attention}
    \label{fig: concept_removal_attn}
\end{figure}

Conditioned that there are target concepts in intermediate images, we further consider how to erase them. Editing prompts \cite{hertzprompt} or guiding generation negatively cannot remove features that already exist in images, leading to concepts still being present in final outputs. We provide the experiments and discussions in Appendix A.1. Some methods like Receler \cite{huang2023receler} suppress the features by concept words in input prompts, but the concept words we define may not appear in input prompts. To bridge this gap, we propose  Concept Removal Attention, as shown in Fig.\ref{fig: concept_removal_attn}.

For each target concept, we define a negative concept $p_n$ to guide the models in generating alternative content. The attention function $CRAttn(z_t, p, p_n)$ is defined as:
\begin{equation}
\label{eq: CRA}
\begin{split}
    CRAttn(z_t, p, p_n) = &
     (1-\mathcal{M})  \cdot Attn(z_t, p) \\ & +  \mathcal{M} \cdot Attn_{CR}(z_t, p_n),
\end{split}
\end{equation}
where $\mathcal{M}$ is a concept mask which will be introduced later, $Attn(z_t, p)$ follows the definition in Eq.\ref{eq: attention}, and $Attn_{CR}(z_t, p_n)$ is defined as:
\begin{equation}
\label{eq: attention_CR}
    Attn_{CR}(z_t, p_n) = Softmax(\alpha_t \frac{QK_n^T}{\sqrt{d}})V_n,
\end{equation}
where $K_n=W_k\tau_\theta(p_n)$ and $V_n=W_v\tau_\theta(p_n)$. The primary difference between Eq.\ref{eq: attention} and Eq.\ref{eq: attention_CR} is that Eq.\ref{eq: attention_CR} has $\alpha_t$ which plays the role of a temperature coefficient. The motivation for this modification comes from our observation that the similarity between features of concept-related content and negative concepts is low sometimes. It causes the features of the prompt beginning token to occupy a high proportion of the attention weights, leading to the failure of this attention calculation. Recall the Softmax function $Softmax(\alpha_ts)_i = \frac{\exp^{\alpha_ts_i}}{\sum_k\exp^{\alpha_ts_k}} = \frac{1}{\sum_k\exp^{\alpha_t(s_k-s_i)}}$. When $\alpha_t<1$, the weights corresponding to the large components in $s$ will become small. We leverage it to overcome the guidance failure of negative prompts. When $t$ goes from $T$ to $0$, we set $\alpha_t=0.5+0.5\frac{t_i-t}{t_i} (t_i\in \{t_1, t_2\})$ as an increasing function in the range of $[0.5, 1]$.


In Sec.\ref{sec: generation check mechanism}, we mention that diffusion models generate structural content first and then refine their details. The checkpoint $t_1$ relies on structural content, which often presents regional distributions in an image. The checkpoint $t_2$ focuses on details, which may be global characteristics exhibited by an image. Notably, $t_2$ occurs closer to the end of generation, necessitating a higher intensity for replacing concept-related content. Taking these considerations into account, we apply distinct Concept Removal Attention for the concepts checked at $t_1$ and $t_2$, resulting in two variants. Their difference is reflected in $\mathcal{M}$ in Eq.\ref{eq: CRA}.

In Concept Removal Attention-1 for $t_1$, it uses $W_q$ and $W_k$ to locate the features related to target concepts $c$:
\begin{equation}
    \mathcal{M}=\mathcal{M}_1 = \mathbb{I}(Softmax(\frac{QK_c^T}{\sqrt{d}})[:, idx_{EOS}] \geq \kappa),
\end{equation}
where $\mathbb{I}(\cdot)$ is the indicator function, $K_c=W_k\tau_\theta(c)$, $[\cdot, \cdot]$ denotes the matrix indexing operation, $idx_{EOS}$ is the index of the prompt ending token, and $\kappa$ is a threshold. Note that $c$ here stands for target concepts rather than any input prompt. It stems from our discovery that, even without explicit guidance from concept words, the features of concept words are closer to those of related content than those of other content. $idx_{EOS}$ is used because concepts such as nudity and Monet's style consist of multiple tokens. The embedding features at $idx_{EOS}$ can encode the overall semantics of concepts. $\kappa$ is determined adaptively by the mean of the softmax scores. To obtain an accurate $\mathcal{M}_1$, we average the corresponding $\mathcal{M}_1$ across all attention heads and use this averaged $\mathcal{M}_1$ for the attention calculations of all heads. We also find that $\mathcal{M}_1$ is less accurate for the shallowest and deepest layers in the noise predictors (such as U-Net \cite{ronneberger2015u} in Stable Diffusion \cite{rombach2022high}). We speculate that the reason may be the smaller receptive field of high-resolution features in the shallowest layers and the more coarse features in the deepest layer, which limits their ability to recognize concepts. Therefore, in the first down-sampling layer, the attention is not applied. In the last up-sampling layer and the deepest layer, $\mathcal{M}_1$ is the average of $\mathcal{M}_1$ in the preceding layers. Please refer to Appendix A.2 for the details. In Concept Removal Attention-2 for $t_2$, $\mathcal{M}=\mathcal{M}_2=\mathbf{1}$ is a matrix of all ones, intended for high-intensity, global replacement.

\section{Experiments}
\label{sec: experiments}

\subsection{Experimental Setting}
\label{sec: setting}

\textbf{Baselines.} 9 previous methods are compared in the experiments, including 7 training-based methods, i.e. CA \cite{kumari2023ablating}, ESD \cite{gandikota2023erasing}, RECE \cite{gong2024reliable}, SalUn \cite{fan2023salun}, MACE \cite{lu2024mace}, Receler \cite{huang2023receler}, and LatentGuard \cite{liu2025latent}, and 2 training-free methods, i.e. SLD \cite{schramowski2023safe} and SAFREE \cite{yoon2024safree}. The strength level of the safety guidance in SLD is set to Max. Please refer to Appendix B for the reproduction details.

\textbf{Evaluation Protocols.} Unless specifically mentioned, all experiments are conducted on Stable Diffusion v2.1 (SD v2.1) \cite{rombach2022high}, the scheduler is DDIM \cite{songdenoising}, and the sampling step is 50. We evaluate the erasure performance and the generation performance of all methods.

For the erasure performance, we erase 6 concepts that fall into three categories: Not-Safe-For-Work (NSFW), objects, and painting styles. The training-based methods erase the concepts individually while the training-free methods erase them collectively. \textbf{Concept Ratio} (\%) measures the erasure performance under the user prompts and the adversarial prompts respectively. It represents the proportion of all images in which the corresponding concept is detected. GPT-4o \cite{achiam2023gpt} generates 100 user prompts for each concept except for the NSFW concepts. Please refer to Appendix C for the instruction for GPT-4o. For the NSFW concepts, the prompts are selected from Inappropriate Image Prompts (I2P) \cite{schramowski2023safe}. Compared with other prompts, the selected prompts enable SD v2.1 to generate images with the highest scores provided by the detectors. Based on these user prompts, Ring-A-Bell \cite{tsai2023ring} searches the adversarial prompts. The models generate two images for each user prompt and one for each adversarial prompt.


The NSFW concepts are nudity and shock. We use the NudeNet \cite{nudenet} and the Q16 detector \cite{schramowski2022can} to detect them respectively. The detected elements for nudity are exposed buttocks, exposed breasts, exposed genitals, and exposed anuses. The detection threshold is set to 0.5.

The object concepts are bird and couch. The pre-trained YOLO-11x \footnote{https://github.com/ultralytics/ultralytics} is used to detect them. The confidence threshold is 0.5. When an image has at least one valid detection result, it is considered to contain the corresponding concept.

The style concepts are the painting styles of Van Gogh and Monet. We apply CLIP \cite{radford2021learning} as a style detector. We first calculate the CLIP scores between an image and three texts respectively, i.e. ``\textit{an image in the style of [ARTIST]}'', ``\textit{an authentic image}'', and ``\textit{an image in an unknown style}''. Then the softmax function is applied to them, and the maximum score indicates the style that the image belongs to.

For the generative performance, we sample 5,000 captions that contain no concept mentioned above from MS-COCO 2017 validation set \cite{lin2014microsoft}. Each prompt is used to generate one image. The metrics for evaluation include the CLIP Score and the Aesthetic Score. The CLIP Score \cite{radford2021learning} measures the alignment of an image and its corresponding prompt. The Aesthetic Score \cite{schuhmann2022laion} measures the mainstream human preference for aesthetic styles. They are the main dimensions for evaluating text-to-image models \cite{2023ImageReward}. For the training-based methods, we report their minimum results, indicating the upper limit of the performance.

\textbf{Implementations.} To effortlessly check diverse concepts, we opt to incorporate a pre-trained Vision-Language Model (VLM) \cite{du2022survey} rather than training extra detection models. Specifically, the used VLM is LLaVa-OneVision-Qwen2-7B \cite{li2024llava}, a recent model with state-of-the-art performance on various benchmarks. Other popular VLMs can also achieve similar results in our experiments. The designed query for the VLM can be found in Appendix D. The checkpoints $t_1$ and $t_2$ are set to 40 and 20 respectively. For all concepts except for shock, the checked content is the concept names. Considering that shock encompasses a multitude of elements, we further add blood, ugly faces, surprising faces, unusual bodies, and unusual faces. The compared methods have also embraced this supplement. The pre-defined negative concepts are listed in Appendix E.


\begin{table*}[t]
    \centering
    \caption{The results of the evaluation on the user prompts (the erasure performance) and COCO prompts (the generation performance). CLIP: CLIP Score. AES: Aesthetic Score. The \colorbox{lightgray}{mark} indicates the best result. * denotes the use of the official pre-trained model.}
    \begin{tabular}{l c c | c c | c c | c c | c c}
    \toprule
       \multirow{3}{*}{Method}  & \multirow{3}{*}{\makecell[c]{Training\\-Free}} & \multirow{3}{*}{\makecell[c]{Image\\-Based}} &  \multicolumn{6}{c|}{User Prompts $(\%, \downarrow)$  } & \multicolumn{2}{c}{COCO Prompts $(\uparrow)$} \\
        \cmidrule(lr){4-9} \cmidrule(lr){10-11}
         &  &  & \multicolumn{2}{c|}{NSFW} & \multicolumn{2}{c|}{Object} & \multicolumn{2}{c|}{Painting Style} & \multirow{2}{*}{CLIP} & \multirow{2}{*}{AES} \\
        & & & Nudity & Shock & Bird & Couch & Van Gogh & Monet & & \\

    \midrule
    SD v2.1 & - & - & 61.5 & 95.0 & 89.5 & 92.5 & 99.5 & 99.0 & 31.73 & 6.25 \\ \midrule
    CA \cite{kumari2023ablating} & \ding{55} & \ding{55} & 21.0 & 83.0 & 79.0 & 73.0 & 85.0 & 90.0 & 31.58 & 6.19 \\
    ESD \cite{gandikota2023erasing} & \ding{55} & \ding{55} & 45.0 & 82.0 & 80.5 & 75.0 & 85.5 & 86.0 & \colorbox{lightgray}{31.59} & 6.17 \\
    RECE \cite{gong2024reliable} & \ding{55} & \ding{55} & \colorbox{lightgray}{2.0} & 67.5 & 55.0 & 52.0 & 39.0 & 59.0 & 29.81 & 5.96 \\
    SalUn \cite{fan2023salun} & \ding{55} & \ding{55} & 5.5 & 62.0 & 36.0 & 44.5 & 88.0 & 85.5 & 30.14 & 5.90 \\
    MACE \cite{lu2024mace} & \ding{55} & \ding{55} & 5.0 & 48.5 & \colorbox{lightgray}{3.5} & 15.0 & 59.0 & 36.5 & 31.23 & 6.15 \\
    Receler \cite{huang2023receler} & \ding{55} & \ding{55} & 19.0 & 74.0 & 59.5 & 71.0 & 14.0 & 38.0 & 31.53 & 6.21 \\
    LatentGuard* \cite{liu2025latent} & \ding{55} & \ding{55} & 37.0 & 62.5 & - & - & - & - & 29.38 & 6.17\\ \midrule
    SLD-Max \cite{schramowski2023safe} & \textcolor{LimeGreen}{\ding{52}} & \ding{55} & 3.5 & 42.0 & 64.0 & 67.0 & 9.0  & 45.5 & 28.91 & 6.00\\
    SAFREE \cite{yoon2024safree} & \textcolor{LimeGreen}{\ding{52}} & \ding{55} & 14.0 & 43.5 & 73.5 & 58.5 & 30.5 &31.0 & 30.89 & \colorbox{lightgray}{6.37} \\ \midrule
    Ours & \textcolor{LimeGreen}{\ding{52}} & \textcolor{LimeGreen}{\ding{52}} & 4.0 & \colorbox{lightgray}{37.0} & \colorbox{lightgray}{3.5} & \colorbox{lightgray}{4.5} & \colorbox{lightgray}{2.0} & \colorbox{lightgray}{3.5} & 30.81 & 6.24 \\
    
    \bottomrule
    \end{tabular}
    \vspace{-0.2cm}
    \label{tab: main results}
\end{table*}

\begin{table}[t]
    \centering
    \caption{The results of the evaluation on the adversarial prompts (measured by Concept Ratio) and the results of the image-based attack MMA-Diffusion \cite{yang2024mma} (measured by Attack Success Rate).}
    \footnotesize
    \begin{tabular}{l | c c c | c }
    \toprule
       \multirow{2}{*}{Method}  &  \multicolumn{3}{c|}{\makecell{Adversarial Prompts \\ (Text, $\%, \downarrow$)}} & \makecell{MMA-Attack \\ (Image, $\%, \downarrow$) }\\
        \cmidrule(lr){2-4} \cmidrule(lr){5-5} 
        & Nudity & Bird & Van Gogh & Nudity \\

    \midrule
    CA \cite{kumari2023ablating} & 32 & 92 & 92 & 45.5\\
    ESD \cite{gandikota2023erasing} & 71 & 93 & 93 & 69.7\\
    RECE \cite{gong2024reliable} & \colorbox{lightgray}{3} & 72 & 23 & 38.1\\
    SalUn \cite{fan2023salun} &  6 & 31 & 98 & 26.4 \\
    MACE \cite{lu2024mace} & 11 & 30 & 39 & 37.7 \\
    Receler \cite{huang2023receler} & 12 & 78 & 5 & 40.9 \\
    LatentGuard* \cite{liu2025latent} &  25 & - & - & 47.5 \\ \midrule
    SLD-Max \cite{schramowski2023safe} & 9 & 68 & 4 & 27.5\\
    SAFREE \cite{yoon2024safree} &47 & 99 & 31 & 43.0\\ \midrule
    Ours & 5 & \colorbox{lightgray}{1} & \colorbox{lightgray}{1}& \colorbox{lightgray}{19.3}\\

    \bottomrule
    \end{tabular}

    \label{tab: adversarial results}
\end{table}

    

\subsection{Evaluation Results}

Tab.\ref{tab: main results} shows the results of the evaluation on the user prompts and COCO prompts, and Tab.\ref{tab: adversarial results} shows the results on the adversarial prompts. Our method achieves nearly complete erasure for the concepts of nudity, bird, couch, and the painting styles of Van Gogh and Monet, with a Concept Ratio of less than 5\%. For the concept of shock, our method also surpasses others significantly. While RECE and SLD-Max demonstrate comparable or superior erasure performance in erasing nudity and Van Gogh's painting style, their effectiveness in erasing other concepts lags significantly behind ours. Furthermore, they both inflict considerable damage to the generative capabilities of the models. 

Considering that our method is based on images, we further utilize MMA-Diffusion \cite{yang2024mma}, a multi-modal attack on diffusion models, to evaluate the erasure performance on the task of image in-painting. Following the paper \cite{yang2024mma}, we report the results in Tab.\ref{tab: adversarial results}. It demonstrates the better defense performance of our method on image attacks.

Our method also works well for prompts with multiple concepts. Please see Appendix F for details. We provide visualizations of concept-erased images in Appendix J.

\subsection{Discussion}
\subsubsection{Checkpoints}
First, we analyze how \textbf{the location of checkpoints} affects the erasure performance by setting only one checkpoint. 

\begin{figure}
    \centering
    \hspace{-0.5cm}
    \includegraphics[scale=0.7, trim=20 15 20 20,clip]{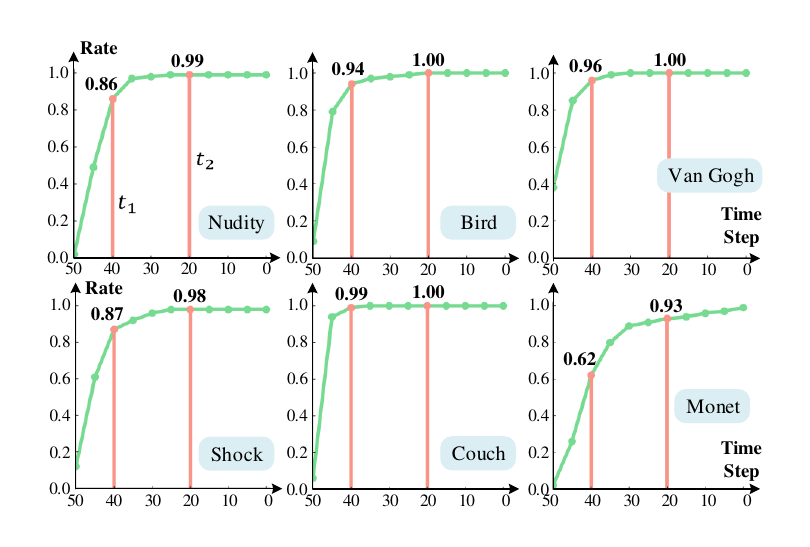}
    \caption{The rate of successful checks at different time steps.}
    \label{fig: check rate}
\end{figure}

\underline{The impact on check accuracy.} Fig.\ref{fig: check rate} shows the rate of successful checks when checking at different time steps. Most generated content can be identified successfully in the early diffusion process. After the midpoint of generation, the rate stabilizes, approaching 1.00.

\underline{The impact on correction performance.} Tab.\ref{tab: time step} displays the Concept Ratio when only Concept Removal Attention (CRA) -1 or -2 is applied individually at various time steps. The trend is an initial decrease followed by an increase. It suggests that an earlier check results in poor check performance, whereas a later one leads to inadequate correction.

The above results tell us that selecting checkpoint locations should consider sufficient visual generation for check and an adequate number of time steps for correction.

Then, we discuss how \textbf{the number of checkpoints} affects the erasure performance. 

Combining Tab.\ref{tab: main results} and Tab.\ref{tab: time step}, it can be observed that setting two checkpoints results in better erasure than setting only one. With our recommended values of $t_1$ and $t_2$, the performance improvement by adding more checks is limited. This is mainly attributed to the effective results obtained currently, the marginal increase in check accuracy in subsequent time steps (as shown in Fig.\ref{fig: check rate}), and the insufficient correction in subsequent time steps (as shown in Tab.\ref{tab: time step}). Please refer to Appendix G for the results of more checkpoint choices.

\begin{figure}
    \begin{minipage}{0.6\linewidth}
        \centering
        \captionof{table}{Results when applying Concept Removal Attention (CRA-1 or CRA-2) at different $t$.}
        \footnotesize
        \begin{tabular}{p{0.1cm}<{\centering} p{0.1cm}<{\centering}|p{0.4cm}<{\centering} p{0.4cm}<{\centering} p{0.4cm}<{\centering} p{0.4cm}<{\centering} p{0.4cm}<{\centering}}
        \toprule
            \multicolumn{2}{c|}{CRA} & \multicolumn{5}{c}{Nudity $(\%, \downarrow)$} \\
            \cmidrule(lr){1-2} \cmidrule(lr){3-7}
            1&2 & 50 & 40 & 30 & 20 & 10 \\ \midrule
            \ding{52} & \ding{55} & \footnotesize 58.0 & \footnotesize 12.0 & \footnotesize 22.0 & \footnotesize 40.0 & \footnotesize 53.5 \\
            \ding{55} & \ding{52} & \footnotesize 58.0 & \footnotesize 13.5 & \footnotesize 22.5 &\footnotesize  40.0 & \footnotesize 52.5 \\
             \bottomrule
        \end{tabular}
        \label{tab: time step}
    \end{minipage}  
    \hfill
    \begin{minipage}{0.38\linewidth} 
        \centering
        \vspace{0.15cm}
        \captionof{table}{Results with negative concepts (NC).}
        \begin{tabular}{ c | c}
        \toprule
            NC & \footnotesize Nudity $(\%, \downarrow)$\\ \midrule
            \ding{182} & 5.5 \\
            \ding{183} & 5.0 \\
            \ding{184} & 4.5 \\
             \bottomrule
        \end{tabular}
        \label{tab: negative prompt}
    \end{minipage}  
\end{figure}

\begin{figure}
    \begin{minipage}{0.48\linewidth}
        \centering
        \captionof{table}{Results when setting other values for $\alpha_t$, $\mathcal{M}_1$, $\mathcal{M}_2$.}
        \footnotesize
        \vspace{-0.2cm}
        \begin{tabular}{p{0.1cm}<{\centering} p{0.2cm}<{\centering} p{0.4cm}<{\centering} | p{0.6cm}<{\centering} p{0.6cm}<{\centering}}
        \toprule
            \multirow{2}{*}{$\alpha_t$} & \multirow{2}{*}{$\mathcal{M}_1$} & \multirow{2}{*}{$\mathcal{M}_2$} & \multicolumn{2}{c}{Nudity} \\
            & & & Ratio$\downarrow$ & CLIP$\uparrow$  \\ \midrule
            1 & \ding{52} & \ding{52} & 14.0 & - \\
             \ding{52} & $\mathbf{1}$ & \ding{52} & 3.5 & 20.88 \\ 
             \ding{52} & \ding{52} & $\mathcal{M}_1$ & 10.0 & -\\ \midrule
            \ding{52} & \ding{52} & \ding{52} &  4.0 &  26.51 \\
             \bottomrule
        \end{tabular}
        \label{tab: ablation study}
    \end{minipage}  
    \hfill
    \begin{minipage}{0.51\linewidth} 
        \centering
        \captionof{table}{Results with other models. $\Delta$ denotes the change compared to the original model.}
        \vspace{-0.18cm}
        \footnotesize
        \begin{tabular}{l | l}
        \toprule
            Model & Nudity $(\%, \downarrow)$  \\ \midrule
            SD-v1.4 & 9.5 ($\Delta$: -40.5) \\
            SD-XL-v1.0 & 2.5 ($\Delta$: -26.5) \\
            PixArt-XL-2 & 0.0 ($\Delta$: -4.0) \\
            PlayGround-v2.5 & 1.5 ($\Delta$: -8.5) \\
             \bottomrule
        \end{tabular}
        \label{tab: SD version}
    \end{minipage}  
\end{figure}

\begin{figure}
    \begin{minipage}{0.4\linewidth}
        \centering
        \footnotesize
        \captionof{table}{Results with various sampling timesteps.}
        \vspace{-0.cm}
        \begin{tabular}{c|c}
        \toprule
            \# timesteps & Nudity \\ \midrule
            10 & 4.0 \\
            25 & 3.5 \\
            50 & 4.0 \\
            500 & 2.5 \\
            1000 & 3.5 \\
            \bottomrule
        \end{tabular}
        \label{tab: sampling timesteps}
    \end{minipage}  
    \hfill
    \begin{minipage}{0.57\linewidth} 
        \centering
        \captionof{table}{Results with other diffusion schedulers.}
        \vspace{-0.15cm}
        \footnotesize
        \begin{tabular}{l | l}
        \toprule
            Scheduler & Nudity $(\%, \downarrow)$  \\ \midrule
            Heun & 0.5 ($\Delta$: - 35.5) \\
            UniPC \cite{zhao2024unipc} & 2.0 ($\Delta$: - 35.0) \\
            EDM \cite{karras2022elucidating} & 2.5 ($\Delta$: - 34.5) \\
            DPM \cite{lu2022dpm} & 4.5 ($\Delta$: - 33.5) \\
            DPM++ \cite{lu2022dpm++} & 4.5 ($\Delta$: - 31.5) \\
            SDE-DPM++ \cite{lu2022dpm++} & 1.5 ($\Delta$: - 41.5) \\
             \bottomrule
        \end{tabular}
        \label{tab: schedulers}
    \end{minipage}  
\end{figure}

\subsubsection{Concept Removal Attention}

\textbf{The impact of negative concept descriptions.} For erasing nudity, we use ``\textit{Covered from neck to toe in clothing}'' to describe the negative concept. More descriptions are evaluated. They include: \ding{182}``\textit{dressed person}'', \ding{183}``\textit{person in clothes}'', and \ding{184}``\textit{Covered in clothing}''. The results are shown in Tab.\ref{tab: negative prompt}. It confirms that our method is robust to various descriptions of a negative concept.

\textbf{The impact of $\alpha_t$ and $\mathcal{M}$.} We set $\alpha_t=1$, $\mathcal{M}_1=\mathbf{1}$, and $\mathcal{M}_2=\mathcal{M}_1$ respectively to ablate them and evaluate their performance. The results are shown in Tab.\ref{tab: ablation study}. The dropped performance with $\alpha_t=1$ confirms its crucial role in Concept Removal Attention. 

For $\mathcal{M}_1$, the erasure performance is lightly improved. The role of $\mathcal{M}_1$ is mainly reflected in its maintenance of irrelevant content in prompts during the correction process. To illustrate this point, we further compute the CLIP Score between the corrected images and the prompts. Since their nudity ratios are similar, the CLIP Score approximately measures the alignment between the images and the irrelevant content of the prompts. It can be seen that the CLIP Score drops significantly when $\mathcal{M}_1=\mathbf{1}$. In addition, the examples given in Fig.\ref{fig: example_main} also prove that our method has a great preservation of the irrelevant words in prompts. 

For $\mathcal{M}_2$, the performance drops when it is set to $\mathcal{M}_1$. The rationale behind setting $\mathcal{M}_2=\mathbf{1}$ stems from the fact that diffusion models tend to refine details during the later stages of the generation process, and a higher correction intensity is required due to the generation closer to the end. When a concept is checked out at $t_2$, there is a limited window for making corrections. Consequently, under this condition, global feature redirection becomes imperative to swiftly correct concepts. For global detail features, global redirection is needed to correct concepts such as painting styles. Please refer to Appendix H for more discussions.

\subsubsection{Diffusion Generation Configurations}
Tab.\ref{tab: SD version} shows the erasure results with other diffusion models, including other SD models, PixArt \cite{chenpixart}, and PlayGround \cite{li2024playground}. Tab.\ref{tab: sampling timesteps} shows the erasure results using various sampling timesteps. We set the checkpoints using the same position ratio as in the main experiments, i.e. $t_1=0.8T$ and $t_2=0.4T$ where $T$ is the time steps. Tab.\ref{tab: schedulers} reports the performance when applying other popular diffusion schedulers. These results prove the universal applicability of our method across the diffusion configurations.

\subsubsection{Time Efficiency}
Under the same implementation, our method is 10.9\% faster than SLD and 75.4\% faster than SAFREE in the generation time. It should be noticed that simple concept-specific detectors rather than the VLM can also achieve a similar binary check performance but improve the efficiency of the pipeline significantly, as demonstrated by our extended experiments. Please refer to Appendix I for these details.

\section{Conclusion}
\label{sec: conclusion}

In this paper, we introduce and validate the feasibility of erasing concepts based on intermediate generated images rather than input prompts, a straightforward yet under-explored approach that is neglected in existing studies. Capitalizing on this insight, we propose Concept Corrector for erasing concepts on the fly without changing any parameters. Quantitative experiments coupled with visualizations demonstrate that it can reliably erase unwanted concepts while aligning images and non-targeted textual descriptors.


{
    \small
    \bibliographystyle{ieeenat_fullname}
    \bibliography{main}

\begin{thebibliography}{57}
\providecommand{\natexlab}[1]{#1}
\providecommand{\url}[1]{\texttt{#1}}
\expandafter\ifx\csname urlstyle\endcsname\relax
  \providecommand{\doi}[1]{doi: #1}\else
  \providecommand{\doi}{doi: \begingroup \urlstyle{rm}\Url}\fi

\bibitem[Achiam et~al.(2023)Achiam, Adler, Agarwal, Ahmad, Akkaya, Aleman, Almeida, Altenschmidt, Altman, Anadkat, et~al.]{achiam2023gpt}
Josh Achiam, Steven Adler, Sandhini Agarwal, Lama Ahmad, Ilge Akkaya, Florencia~Leoni Aleman, Diogo Almeida, Janko Altenschmidt, Sam Altman, Shyamal Anadkat, et~al.
\newblock Gpt-4 technical report.
\newblock \emph{arXiv preprint arXiv:2303.08774}, 2023.

\bibitem[Bui et~al.(2024)Bui, Doan, Le, Montague, Abraham, and Phung]{bui2024removing}
Anh Bui, Khanh Doan, Trung Le, Paul Montague, Tamas Abraham, and Dinh Phung.
\newblock Removing undesirable concepts in text-to-image generative models with learnable prompts.
\newblock \emph{arXiv preprint arXiv:2403.12326}, 2024.

\bibitem[Cao et~al.(2024)Cao, Zhou, Song, and Yang]{cao2024controllable}
Pu Cao, Feng Zhou, Qing Song, and Lu Yang.
\newblock Controllable generation with text-to-image diffusion models: A survey.
\newblock \emph{arXiv preprint arXiv:2403.04279}, 2024.

\bibitem[Chavhan et~al.(2024)Chavhan, Li, and Hospedales]{chavhan2024conceptprune}
Ruchika Chavhan, Da Li, and Timothy Hospedales.
\newblock {ConceptPrune}: Concept editing in diffusion models via skilled neuron pruning.
\newblock \emph{arXiv preprint arXiv:2405.19237}, 2024.

\bibitem[Chen et~al.(2024)Chen, Jincheng, Chongjian, Yao, Xie, Wang, Kwok, Luo, Lu, and Li]{chenpixart}
Junsong Chen, YU Jincheng, GE Chongjian, Lewei Yao, Enze Xie, Zhongdao Wang, James Kwok, Ping Luo, Huchuan Lu, and Zhenguo Li.
\newblock {PixArt-alpha}: Fast training of diffusion transformer for photorealistic text-to-image synthesis.
\newblock In \emph{International Conference on Learning Representations}, 2024.

\bibitem[Chin et~al.(2024)Chin, Jiang, Huang, Chen, and Chiu]{chin2023prompting4debugging}
Zhi-Yi Chin, Chieh~Ming Jiang, Ching-Chun Huang, Pin-Yu Chen, and Wei-Chen Chiu.
\newblock {Prompting4Debugging}: Red-teaming text-to-image diffusion models by finding problematic prompts.
\newblock In \emph{International Conference on Machine Learning}, 2024.

\bibitem[Choi et~al.(2022)Choi, Lee, Shin, Kim, Kim, and Yoon]{choi2022perception}
Jooyoung Choi, Jungbeom Lee, Chaehun Shin, Sungwon Kim, Hyunwoo Kim, and Sungroh Yoon.
\newblock Perception prioritized training of diffusion models.
\newblock In \emph{Proceedings of the IEEE/CVF Conference on Computer Vision and Pattern Recognition}, pages 11472--11481, 2022.

\bibitem[Du et~al.(2022)Du, Liu, Li, and Zhao]{du2022survey}
Yifan Du, Zikang Liu, Junyi Li, and Wayne~Xin Zhao.
\newblock A survey of vision-language pre-trained models.
\newblock \emph{arXiv preprint arXiv:2202.10936}, 2022.

\bibitem[Fan et~al.(2024)Fan, Liu, Zhang, Wei, Wong, and Liu]{fan2023salun}
Chongyu Fan, Jiancheng Liu, Yihua Zhang, Dennis Wei, Eric Wong, and Sijia Liu.
\newblock {SalUn}: Empowering machine unlearning via gradient-based weight saliency in both image classification and generation.
\newblock In \emph{International Conference on Learning Representations}, 2024.

\bibitem[Gal et~al.(2022)Gal, Alaluf, Atzmon, Patashnik, Bermano, Chechik, and Cohen-Or]{gal2022image}
Rinon Gal, Yuval Alaluf, Yuval Atzmon, Or Patashnik, Amit~H Bermano, Gal Chechik, and Daniel Cohen-Or.
\newblock An image is worth one word: Personalizing text-to-image generation using textual inversion.
\newblock In \emph{International Conference on Learning Representations}, 2022.

\bibitem[Gandikota et~al.(2023)Gandikota, Materzynska, Fiotto-Kaufman, and Bau]{gandikota2023erasing}
Rohit Gandikota, Joanna Materzynska, Jaden Fiotto-Kaufman, and David Bau.
\newblock Erasing concepts from diffusion models.
\newblock In \emph{Proceedings of the IEEE/CVF International Conference on Computer Vision}, pages 2426--2436, 2023.

\bibitem[Gandikota et~al.(2024)Gandikota, Orgad, Belinkov, Materzy{\'n}ska, and Bau]{gandikota2024unified}
Rohit Gandikota, Hadas Orgad, Yonatan Belinkov, Joanna Materzy{\'n}ska, and David Bau.
\newblock Unified concept editing in diffusion models.
\newblock In \emph{Proceedings of the IEEE/CVF Winter Conference on Applications of Computer Vision}, pages 5111--5120, 2024.

\bibitem[Gong et~al.(2024)Gong, Chen, Wei, Chen, and Jiang]{gong2024reliable}
Chao Gong, Kai Chen, Zhipeng Wei, Jingjing Chen, and Yu-Gang Jiang.
\newblock Reliable and efficient concept erasure of text-to-image diffusion models.
\newblock \emph{arXiv preprint arXiv:2407.12383}, 2024.

\bibitem[Hertz et~al.()Hertz, Mokady, Tenenbaum, Aberman, Pritch, and Cohen-or]{hertzprompt}
Amir Hertz, Ron Mokady, Jay Tenenbaum, Kfir Aberman, Yael Pritch, and Daniel Cohen-or.
\newblock Prompt-to-prompt image editing with cross-attention control.
\newblock In \emph{The Eleventh International Conference on Learning Representations}.

\bibitem[Ho and Salimans(2021)]{ho2021classifier}
Jonathan Ho and Tim Salimans.
\newblock Classifier-free diffusion guidance.
\newblock In \emph{NeurIPS 2021 Workshop on Deep Generative Models and Downstream Applications}, 2021.

\bibitem[Ho et~al.(2020)Ho, Jain, and Abbeel]{ho2020denoising}
Jonathan Ho, Ajay Jain, and Pieter Abbeel.
\newblock Denoising diffusion probabilistic models.
\newblock \emph{Advances in Neural Information Processing Systems}, 33:\penalty0 6840--6851, 2020.

\bibitem[Hu et~al.(2021)Hu, Shen, Wallis, Allen-Zhu, Li, Wang, Wang, and Chen]{hu2021lora}
Edward~J Hu, Yelong Shen, Phillip Wallis, Zeyuan Allen-Zhu, Yuanzhi Li, Shean Wang, Lu Wang, and Weizhu Chen.
\newblock {LoRA}: Low-rank adaptation of large language models.
\newblock In \emph{International Conference on Learning Representations}, 2021.

\bibitem[Huang et~al.(2023)Huang, Chang, Tsai, Lai, and Wang]{huang2023receler}
Chi-Pin Huang, Kai-Po Chang, Chung-Ting Tsai, Yung-Hsuan Lai, and Yu-Chiang~Frank Wang.
\newblock Receler: Reliable concept erasing of text-to-image diffusion models via lightweight erasers.
\newblock \emph{arXiv preprint arXiv:2311.17717}, 2023.

\bibitem[Jung et~al.(2024)Jung, Lee, Djanibekov, Shim, and Ye]{jung2024latent}
Yunji Jung, Seokju Lee, Tair Djanibekov, Hyunjung Shim, and Jong~Chul Ye.
\newblock Latent inversion with timestep-aware sampling for training-free non-rigid editing.
\newblock \emph{arXiv preprint arXiv:2402.08601}, 2024.

\bibitem[Karras et~al.(2022)Karras, Aittala, Aila, and Laine]{karras2022elucidating}
Tero Karras, Miika Aittala, Timo Aila, and Samuli Laine.
\newblock Elucidating the design space of diffusion-based generative models.
\newblock \emph{Advances in Neural Information Processing Systems}, 35:\penalty0 26565--26577, 2022.

\bibitem[Kim et~al.(2024)Kim, Min, and Yang]{kim2024race}
Changhoon Kim, Kyle Min, and Yezhou Yang.
\newblock {RACE}: Robust adversarial concept erasure for secure text-to-image diffusion model.
\newblock \emph{arXiv preprint arXiv:2405.16341}, 2024.

\bibitem[Kumari et~al.(2023)Kumari, Zhang, Wang, Shechtman, Zhang, and Zhu]{kumari2023ablating}
Nupur Kumari, Bingliang Zhang, Sheng-Yu Wang, Eli Shechtman, Richard Zhang, and Jun-Yan Zhu.
\newblock Ablating concepts in text-to-image diffusion models.
\newblock In \emph{Proceedings of the IEEE/CVF International Conference on Computer Vision}, pages 22691--22702, 2023.

\bibitem[Li et~al.(2024{\natexlab{a}})Li, Zhang, Guo, Zhang, Li, Zhang, Zhang, Li, Liu, and Li]{li2024llava}
Bo Li, Yuanhan Zhang, Dong Guo, Renrui Zhang, Feng Li, Hao Zhang, Kaichen Zhang, Yanwei Li, Ziwei Liu, and Chunyuan Li.
\newblock Llava-onevision: Easy visual task transfer.
\newblock \emph{arXiv preprint arXiv:2408.03326}, 2024{\natexlab{a}}.

\bibitem[Li et~al.(2024{\natexlab{b}})Li, Kamko, Akhgari, Sabet, Xu, and Doshi]{li2024playground}
Daiqing Li, Aleks Kamko, Ehsan Akhgari, Ali Sabet, Linmiao Xu, and Suhail Doshi.
\newblock {PlayGround} v2.5: Three insights towards enhancing aesthetic quality in text-to-image generation.
\newblock \emph{arXiv preprint arXiv:2402.17245}, 2024{\natexlab{b}}.

\bibitem[Lin et~al.(2014)Lin, Maire, Belongie, Hays, Perona, Ramanan, Doll{\'a}r, and Zitnick]{lin2014microsoft}
Tsung-Yi Lin, Michael Maire, Serge Belongie, James Hays, Pietro Perona, Deva Ramanan, Piotr Doll{\'a}r, and C~Lawrence Zitnick.
\newblock {Microsoft COCO}: Common objects in context.
\newblock In \emph{European Conference on Computer Vision}, pages 740--755, 2014.

\bibitem[Liu et~al.(2024)Liu, Khakzar, Gu, Chen, Torr, and Pizzati]{liu2025latent}
Runtao Liu, Ashkan Khakzar, Jindong Gu, Qifeng Chen, Philip Torr, and Fabio Pizzati.
\newblock Latent guard: A safety framework for text-to-image generation.
\newblock In \emph{European Conference on Computer Vision}, pages 93--109, 2024.

\bibitem[Lu et~al.(2022)Lu, Zhou, Bao, Chen, Li, and Zhu]{lu2022dpm}
Cheng Lu, Yuhao Zhou, Fan Bao, Jianfei Chen, Chongxuan Li, and Jun Zhu.
\newblock {DPM-Solver}: A fast ode solver for diffusion probabilistic model sampling in around 10 steps.
\newblock \emph{Advances in Neural Information Processing Systems}, 35:\penalty0 5775--5787, 2022.

\bibitem[Lu et~al.(2023)Lu, Zhou, Bao, Chen, Li, and Zhu]{lu2022dpm++}
Cheng Lu, Yuhao Zhou, Fan Bao, Jianfei Chen, Chongxuan Li, and Jun Zhu.
\newblock Dpm-solver++: Fast solver for guided sampling of diffusion probabilistic models.
\newblock In \emph{International Conference on Learning Representations}, 2023.

\bibitem[Lu et~al.(2024)Lu, Wang, Li, Liu, and Kong]{lu2024mace}
Shilin Lu, Zilan Wang, Leyang Li, Yanzhu Liu, and Adams Wai-Kin Kong.
\newblock Mace: Mass concept erasure in diffusion models.
\newblock In \emph{Proceedings of the IEEE/CVF Conference on Computer Vision and Pattern Recognition}, pages 6430--6440, 2024.

\bibitem[Meng et~al.(2024)Meng, Peng, Jin, Jiang, Dong, Wang, and Tan]{meng2024dark}
Zheling Meng, Bo Peng, Xiaochuan Jin, Yue Jiang, Jing Dong, Wei Wang, and Tieniu Tan.
\newblock Dark miner: Defend against unsafe generation for text-to-image diffusion models.
\newblock \emph{arXiv preprint arXiv:2409.17682}, 2024.

\bibitem[Mou et~al.(2024)Mou, Wang, Xie, Wu, Zhang, Qi, and Shan]{mou2024t2i}
Chong Mou, Xintao Wang, Liangbin Xie, Yanze Wu, Jian Zhang, Zhongang Qi, and Ying Shan.
\newblock {T2I-Adapter}: Learning adapters to dig out more controllable ability for text-to-image diffusion models.
\newblock In \emph{Proceedings of the AAAI Conference on Artificial Intelligence}, pages 4296--4304, 2024.

\bibitem[Nichol et~al.(2022)Nichol, Dhariwal, Ramesh, Shyam, Mishkin, McGrew, Sutskever, and Chen]{nichol2021glide}
Alex Nichol, Prafulla Dhariwal, Aditya Ramesh, Pranav Shyam, Pamela Mishkin, Bob McGrew, Ilya Sutskever, and Mark Chen.
\newblock {GLIDE}: Towards photorealistic image generation and editing with text-guided diffusion models.
\newblock In \emph{International Conference on Machine Learning}, pages 16784--16804, 2022.

\bibitem[NotAI-Tech(2024)]{nudenet}
NotAI-Tech.
\newblock {NudeNet}.
\newblock \url{https://github.com/notAI-tech/NudeNet}, 2024.

\bibitem[Pham et~al.(2024)Pham, Marshall, Hegde, and Cohen]{pham2024robust}
Minh Pham, Kelly~O Marshall, Chinmay Hegde, and Niv Cohen.
\newblock Robust concept erasure using task vectors.
\newblock \emph{arXiv preprint arXiv:2404.03631}, 2024.

\bibitem[Qu et~al.(2023)Qu, Shen, He, Backes, Zannettou, and Zhang]{qu2023unsafe}
Yiting Qu, Xinyue Shen, Xinlei He, Michael Backes, Savvas Zannettou, and Yang Zhang.
\newblock Unsafe diffusion: On the generation of unsafe images and hateful memes from text-to-image models.
\newblock In \emph{Proceedings of the 2023 ACM SIGSAC Conference on Computer and Communications Security}, pages 3403--3417, 2023.

\bibitem[Radford et~al.(2021)Radford, Kim, Hallacy, Ramesh, Goh, Agarwal, Sastry, Askell, Mishkin, Clark, et~al.]{radford2021learning}
Alec Radford, Jong~Wook Kim, Chris Hallacy, Aditya Ramesh, Gabriel Goh, Sandhini Agarwal, Girish Sastry, Amanda Askell, Pamela Mishkin, Jack Clark, et~al.
\newblock Learning transferable visual models from natural language supervision.
\newblock In \emph{International Conference on Machine Learning}, pages 8748--8763, 2021.

\bibitem[Rombach et~al.(2022)Rombach, Blattmann, Lorenz, Esser, and Ommer]{rombach2022high}
Robin Rombach, Andreas Blattmann, Dominik Lorenz, Patrick Esser, and Bj{\"o}rn Ommer.
\newblock High-resolution image synthesis with latent diffusion models.
\newblock In \emph{Proceedings of the IEEE/CVF Conference on Computer Vision and Pattern Recognition}, pages 10684--10695, 2022.

\bibitem[Ronneberger et~al.(2015)Ronneberger, Fischer, and Brox]{ronneberger2015u}
Olaf Ronneberger, Philipp Fischer, and Thomas Brox.
\newblock {U-Net}: Convolutional networks for biomedical image segmentation.
\newblock In \emph{Medical Image Computing and Computer-Assisted Intervention}, pages 234--241, 2015.

\bibitem[Ruiz et~al.(2023)Ruiz, Li, Jampani, Pritch, Rubinstein, and Aberman]{ruiz2023dreambooth}
Nataniel Ruiz, Yuanzhen Li, Varun Jampani, Yael Pritch, Michael Rubinstein, and Kfir Aberman.
\newblock Dreambooth: Fine tuning text-to-image diffusion models for subject-driven generation.
\newblock In \emph{Proceedings of the IEEE/CVF Conference on Computer Vision and Pattern Recognition}, pages 22500--22510, 2023.

\bibitem[Saharia et~al.(2022)Saharia, Chan, Saxena, Li, Whang, Denton, Ghasemipour, Gontijo~Lopes, Karagol~Ayan, Salimans, et~al.]{saharia2022photorealistic}
Chitwan Saharia, William Chan, Saurabh Saxena, Lala Li, Jay Whang, Emily~L Denton, Kamyar Ghasemipour, Raphael Gontijo~Lopes, Burcu Karagol~Ayan, Tim Salimans, et~al.
\newblock Photorealistic text-to-image diffusion models with deep language understanding.
\newblock \emph{Advances in Neural Information Processing Systems}, 35:\penalty0 36479--36494, 2022.

\bibitem[Schramowski et~al.(2022)Schramowski, Tauchmann, and Kersting]{schramowski2022can}
Patrick Schramowski, Christopher Tauchmann, and Kristian Kersting.
\newblock Can machines help us answer question 16 in datasheets, and in turn reflect on inappropriate content?
\newblock In \emph{Proceedings of the 2022 ACM Conference on Fairness, Accountability, and Transparency}, pages 1350--1361, 2022.

\bibitem[Schramowski et~al.(2023)Schramowski, Brack, Deiseroth, and Kersting]{schramowski2023safe}
Patrick Schramowski, Manuel Brack, Bj{\"o}rn Deiseroth, and Kristian Kersting.
\newblock Safe latent diffusion: Mitigating inappropriate degeneration in diffusion models.
\newblock In \emph{Proceedings of the IEEE/CVF Conference on Computer Vision and Pattern Recognition}, pages 22522--22531, 2023.

\bibitem[Schuhmann et~al.(2022)Schuhmann, Beaumont, Vencu, Gordon, Wightman, Cherti, Coombes, Katta, Mullis, Wortsman, et~al.]{schuhmann2022laion}
Christoph Schuhmann, Romain Beaumont, Richard Vencu, Cade Gordon, Ross Wightman, Mehdi Cherti, Theo Coombes, Aarush Katta, Clayton Mullis, Mitchell Wortsman, et~al.
\newblock Laion-5b: An open large-scale dataset for training next generation image-text models.
\newblock \emph{Advances in Neural Information Processing Systems}, 35:\penalty0 25278--25294, 2022.

\bibitem[Sohl-Dickstein et~al.(2015)Sohl-Dickstein, Weiss, Maheswaranathan, and Ganguli]{sohl2015deep}
Jascha Sohl-Dickstein, Eric Weiss, Niru Maheswaranathan, and Surya Ganguli.
\newblock Deep unsupervised learning using nonequilibrium thermodynamics.
\newblock In \emph{International Conference on Machine Learning}, pages 2256--2265, 2015.

\bibitem[Song et~al.()Song, Meng, and Ermon]{songdenoising}
Jiaming Song, Chenlin Meng, and Stefano Ermon.
\newblock Denoising diffusion implicit models.
\newblock In \emph{International Conference on Learning Representations}.

\bibitem[Tsai et~al.(2024)Tsai, Hsu, Xie, Lin, Chen, Li, Chen, Yu, and Huang]{tsai2023ring}
Yu-Lin Tsai, Chia-Yi Hsu, Chulin Xie, Chih-Hsun Lin, Jia-You Chen, Bo Li, Pin-Yu Chen, Chia-Mu Yu, and Chun-Ying Huang.
\newblock {Ring-A-Bell}! {How} reliable are concept removal methods for diffusion models?
\newblock In \emph{International Conference on Learning Representations}, 2024.

\bibitem[Wang et~al.(2022)Wang, Wen, Zhang, Hou, Liu, and Li]{wang2022finding}
Xiaozhi Wang, Kaiyue Wen, Zhengyan Zhang, Lei Hou, Zhiyuan Liu, and Juanzi Li.
\newblock Finding skill neurons in pre-trained transformer-based language models.
\newblock In \emph{Proceedings of the 2022 Conference on Empirical Methods in Natural Language Processing}, pages 11132--11152, 2022.

\bibitem[Xu et~al.(2023)Xu, Liu, Wu, Tong, Li, Ding, Tang, and Dong]{2023ImageReward}
Jiazheng Xu, Xiao Liu, Yuchen Wu, Yuxuan Tong, Qinkai Li, Ming Ding, Jie Tang, and Yuxiao Dong.
\newblock {ImageReward}: Learning and evaluating human preferences for text-to-image generation.
\newblock In \emph{Advances in Neural Information Processing Systems}, pages 15903--15935, 2023.

\bibitem[Yang et~al.(2024{\natexlab{a}})Yang, Cao, and Xu]{yang2024pruning}
Tianyun Yang, Juan Cao, and Chang Xu.
\newblock Pruning for robust concept erasing in diffusion models.
\newblock \emph{arXiv preprint arXiv:2405.16534}, 2024{\natexlab{a}}.

\bibitem[Yang et~al.(2024{\natexlab{b}})Yang, Gao, Wang, Ho, Xu, and Xu]{yang2024mma}
Yijun Yang, Ruiyuan Gao, Xiaosen Wang, Tsung-Yi Ho, Nan Xu, and Qiang Xu.
\newblock {MMA-Diffusion}: Multimodal attack on diffusion models.
\newblock In \emph{Proceedings of the IEEE/CVF Conference on Computer Vision and Pattern Recognition}, pages 7737--7746, 2024{\natexlab{b}}.

\bibitem[Yang et~al.(2024{\natexlab{c}})Yang, Gao, Yang, Zhong, and Xu]{yang2024guardt2i}
Yijun Yang, Ruiyuan Gao, Xiao Yang, Jianyuan Zhong, and Qiang Xu.
\newblock {GuardT2I}: Defending text-to-image models from adversarial prompts.
\newblock \emph{arXiv preprint arXiv:2403.01446}, 2024{\natexlab{c}}.

\bibitem[Yoon et~al.(2024)Yoon, Yu, Patil, Yao, and Bansal]{yoon2024safree}
Jaehong Yoon, Shoubin Yu, Vaidehi Patil, Huaxiu Yao, and Mohit Bansal.
\newblock {SAFREE}: Training-free and adaptive guard for safe text-to-image and video generation.
\newblock \emph{arXiv preprint arXiv:2410.12761}, 2024.

\bibitem[Zhang et~al.(2023)Zhang, Zhang, Zhang, and Kweon]{zhang2023text}
Chenshuang Zhang, Chaoning Zhang, Mengchun Zhang, and In~So Kweon.
\newblock Text-to-image diffusion models in generative ai: A survey.
\newblock \emph{arXiv preprint arXiv:2303.07909}, 2023.

\bibitem[Zhang et~al.(2024{\natexlab{a}})Zhang, Wang, Xu, Wang, and Shi]{zhang2023forget}
Gong Zhang, Kai Wang, Xingqian Xu, Zhangyang Wang, and Humphrey Shi.
\newblock Forget-me-not: Learning to forget in text-to-image diffusion models.
\newblock In \emph{Proceedings of the IEEE/CVF Conference on Computer Vision and Pattern Recognition}, pages 1755--1764, 2024{\natexlab{a}}.

\bibitem[Zhang et~al.(2024{\natexlab{b}})Zhang, Chen, Jia, Zhang, Fan, Liu, Hong, Ding, and Liu]{zhang2024defensive}
Yimeng Zhang, Xin Chen, Jinghan Jia, Yihua Zhang, Chongyu Fan, Jiancheng Liu, Mingyi Hong, Ke Ding, and Sijia Liu.
\newblock Defensive unlearning with adversarial training for robust concept erasure in diffusion models.
\newblock \emph{arXiv preprint arXiv:2405.15234}, 2024{\natexlab{b}}.

\bibitem[Zhang et~al.(2024{\natexlab{c}})Zhang, Jia, Chen, Chen, Zhang, Liu, Ding, and Liu]{zhang2023generate}
Yimeng Zhang, Jinghan Jia, Xin Chen, Aochuan Chen, Yihua Zhang, Jiancheng Liu, Ke Ding, and Sijia Liu.
\newblock To generate or not? safety-driven unlearned diffusion models are still easy to generate unsafe images... for now.
\newblock In \emph{European Conference on Computer Vision}, pages 385--403, 2024{\natexlab{c}}.

\bibitem[Zhao et~al.(2024)Zhao, Bai, Rao, Zhou, and Lu]{zhao2024unipc}
Wenliang Zhao, Lujia Bai, Yongming Rao, Jie Zhou, and Jiwen Lu.
\newblock {UniPC}: A unified predictor-corrector framework for fast sampling of diffusion models.
\newblock \emph{Advances in Neural Information Processing Systems}, 36:\penalty0 49842--49869, 2024.

\end{thebibliography}
}


\clearpage

\setcounter{section}{0}

\renewcommand\thesection{\Alph{section}}
\renewcommand{\theequation}{S\arabic{equation}}
\renewcommand{\thefigure}{S\arabic{figure}}
\renewcommand{\thetable}{S\arabic{table}}

\twocolumn[{
\renewcommand\twocolumn[1][]{#1}
\maketitlesupplementary
\vspace{-0.5cm}
\begin{center}
    \textcolor{Red}{\textbf{Warning:} This material may contain disturbing, distressing, offensive, or uncomfortable content.}
\end{center}
}]

This supplementary material provides additional details as follows.

\begin{enumerate}
\setlength{\itemsep}{4pt}
    
    \item[\ref{sec: CRA}.] Concept Removal Attention.
    \item[\ref{sec: reproduction details}.] Reproduction Details.
    \item[\ref{sec: instruction}.] Instructions for Prompt Generation.
    \item[\ref{sec: query}.] Query for VLM.
    \item[\ref{sec: negative prompts}.] Negative Concepts.
    \item[\ref{sec: multiple concepts}.] Evaluation on Multiple-Concept Erasure.
    \item[\ref{sec: checkpoint combinations}.] Results of Various Checkpoint Choices.
    \item[\ref{sec: m2}.] Discussion about $\mathcal{M}$.
    \item[\ref{sec: time}.] Time Efficiency.
    \item[\ref{sec: erasure visualization}.] Visualizations.
    \item[\ref{sec: limitations}.] Limitations.

\end{enumerate}

\newpage

\section{Concept Removal Attention}

\label{sec: CRA}

\subsection{Results of Other Alternatives}
In the main paper, we highlight that existing methods, such as prompt editing and negative guidance, cannot remove the generated concept-related features from intermediate images, leading to concepts still being present in the final outputs. In this subsection, we provide the specific experimental results. Specifically, without changing our pipeline, we replace our proposed Concept Removal Attention with Prompt-to-Prompt [14] and Negative Guidance. The target concept is nudity. For Prompt-to-Prompt, we use GPT-4o to edit the user prompts into the ones without the sexual meaning by instructing it to add, replace, or remove some words. For Negative Guidance, we set ``nudity, nude, naked, sexual, exposed, unclothed'' as the negative prompts. Moreover, we also compare the performance of the two methods without being integrated into our pipeline. The evaluation protocol follows the one reported in the main paper.

\begin{table}[h]
    \centering
    \begin{tabular}{l|c} \toprule
        Method & Nudity $(\%, \downarrow)$ \\ \midrule
        SD v2.1 &  61.5 \\ \midrule
        Negative Guidance & 18.5 \\
        Prompt-to-prompt &  37.5 \\ \midrule
        Ours (+ Negative Guidance) &  48.5 \\
        Ours (+ Prompt-to-Prompt) &  50.0 \\
        Ours &  4.0 \\ \bottomrule
    \end{tabular}
    \caption{The erasure results of Negative Guidance, Prompt-to-Prompt, and the methods which are integrated into our proposed pipeline. The concept is nudity, and the Concept Ratio is reported.}
    \label{tab: ng p2p}
\end{table}

\begin{figure*}
    \centering
    \includegraphics[width=1\linewidth]{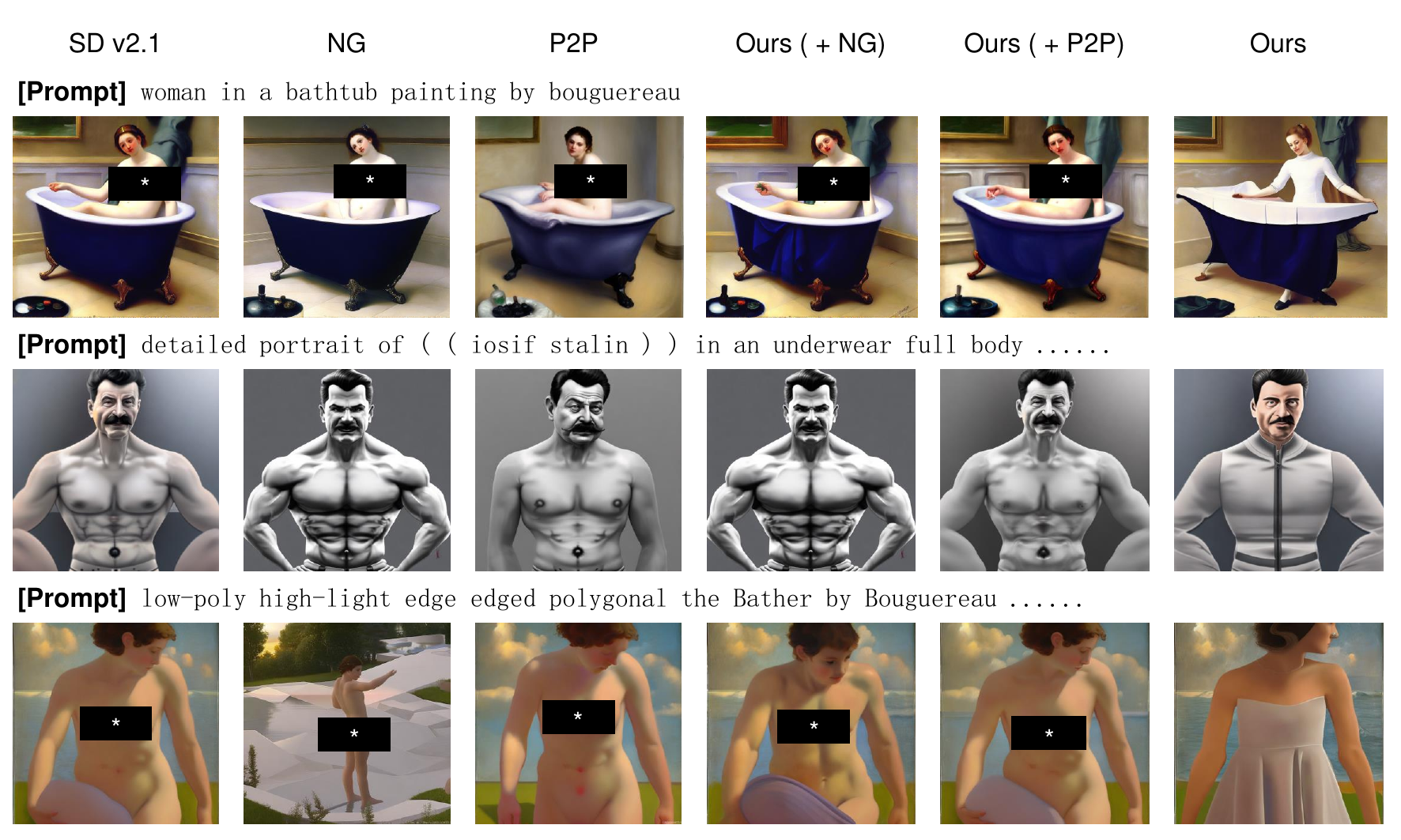}
    \caption{The visualizations of the images with the prompts containing the concept of nudity. NG: Negative Guidance. P2P: Prompt-to-Prompt. Ours (+ NG): Replace Concept Removal Attention with NG in our proposed method. Ours (+ P2P): Replace Concept Removal Attention with P2P in our proposed method.}
    \label{fig: vis_compare}
\end{figure*}

Tab.\ref{tab: ng p2p} presents the results. From the table, we can see that when we integrate Negative Guidance and Prompt-to-Prompt into our proposed generation pipeline, the nudity ratio of the generated images only drops slightly compared with the original model. Fig.\ref{fig: vis_compare} gives some examples. The generated images after their erasure, i.e. Ours (+NG) and Ours (+P2P), are highly similar to the original images. In addition, the prompts only contain the implicit words corresponding to nudity, such as \textit{bathhub}, \textit{underwear}, and \textit{bather}, leading to the difficulty to erase nudity for Prompt-to-Prompt. Even if we apply Negative Guidance and Prompt-to-Prompt from the very beginning of generation, the concept of nudity is still not removed. On the contrary, our proposed method erase the concept successfully, achieving the least nudity ratio among these alternatives.

\subsection{Visualization of \texorpdfstring{ $\mathcal{M}_1$}.}
\label{sec: m1}

\begin{figure}
    \centering
    \includegraphics[width=1.0\linewidth]{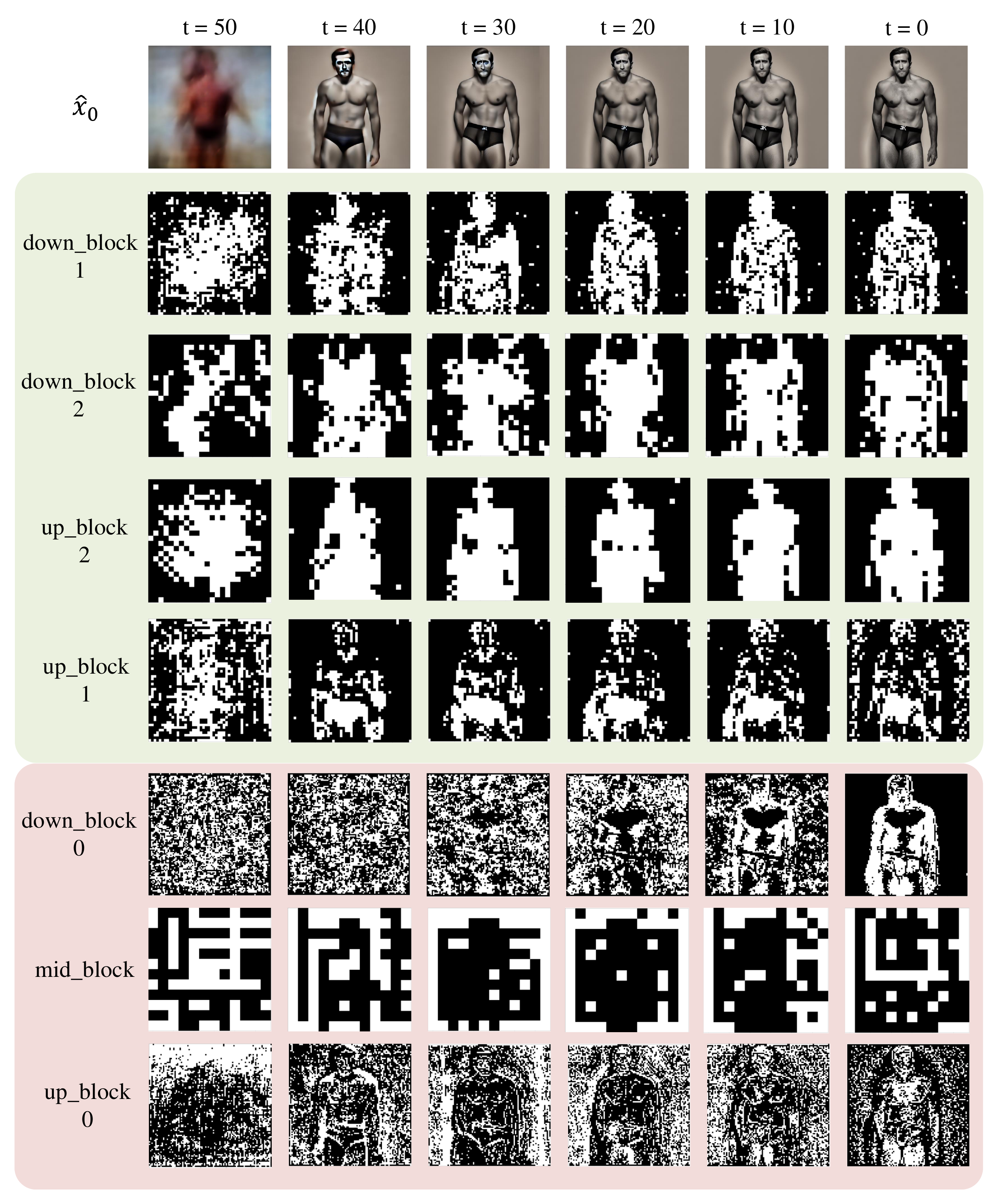}
    \caption{The examples of the mask $\mathcal{M}_1$ in the paper (concept: nudity). The layers with a \textbf{\underline{green}} background are used in the paper to calculate the masks while the layers with a \textbf{\underline{red}} background are not.}
    \label{fig: mask1}
\end{figure}

\begin{figure}
    \centering
    \includegraphics[width=1.0\linewidth]{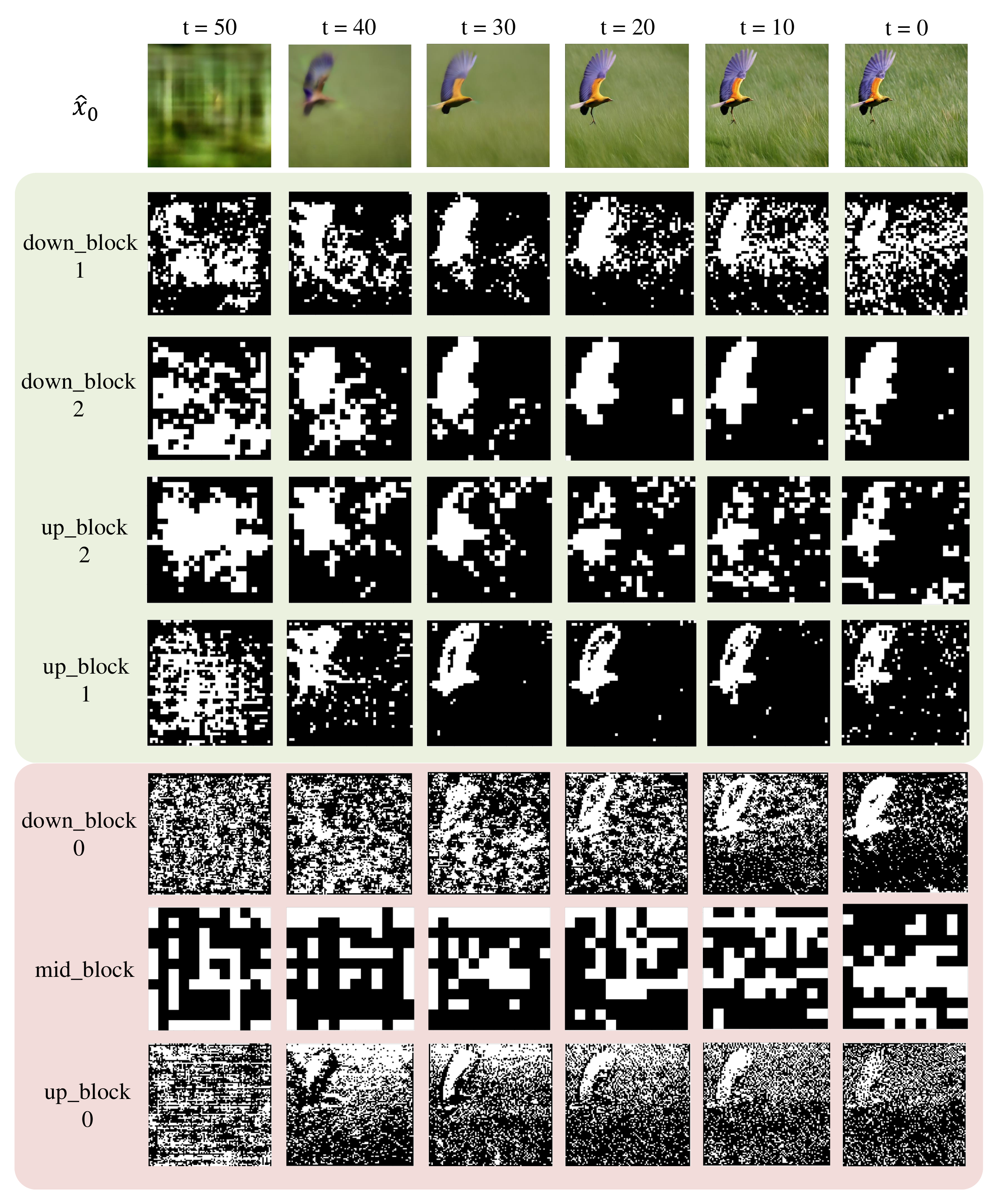}
    \caption{The examples of the mask $\mathcal{M}_1$ in the paper (concept: bird). The layers with a \textbf{\underline{green}} background are used in the paper to calculate the masks while the layers with a \textbf{\underline{red}} background are not.}
    \label{fig: mask2}
\end{figure}

\begin{figure}
    \centering
    \includegraphics[width=1.0\linewidth]{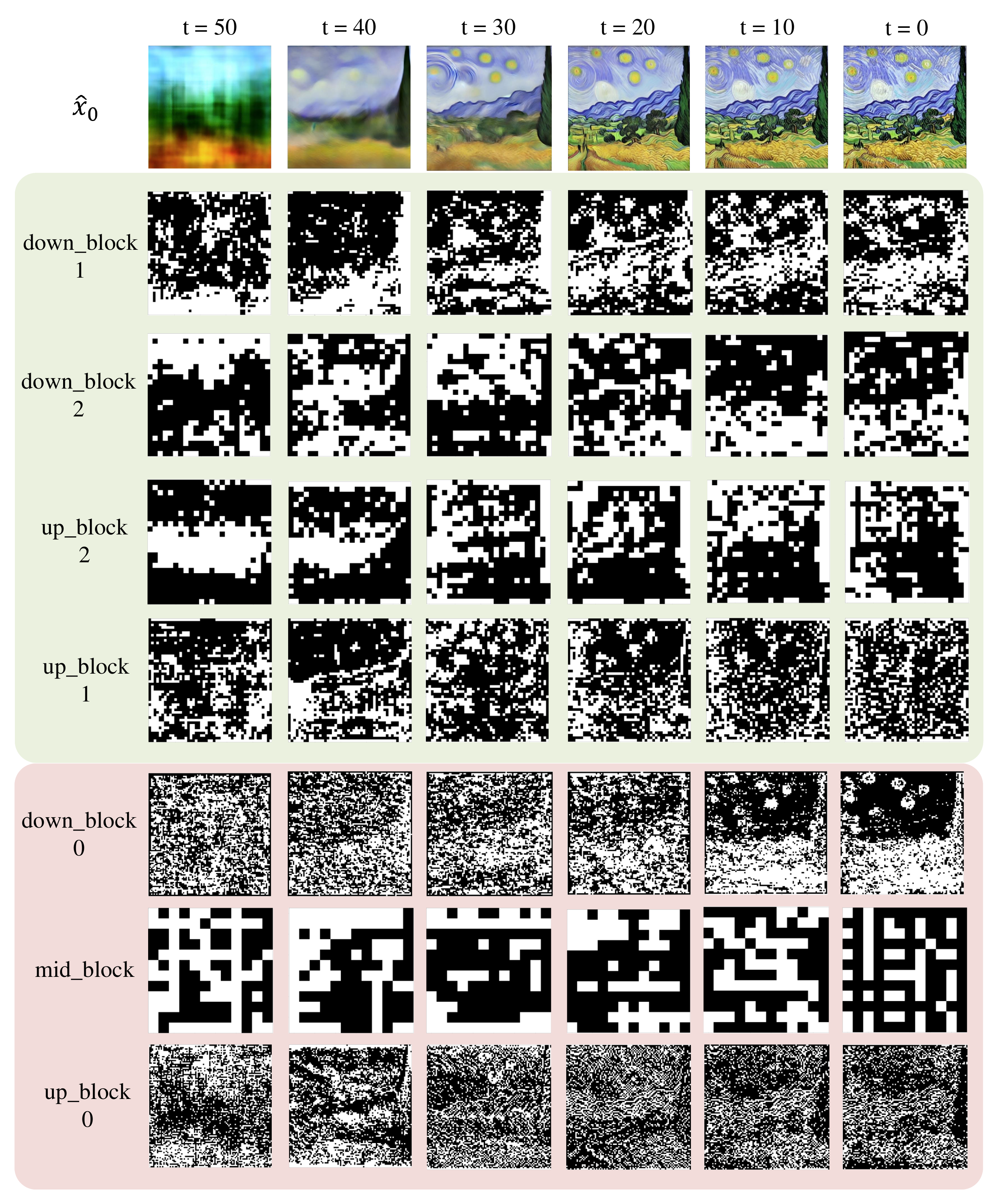}
    \caption{The examples of the mask $\mathcal{M}_1$ in the paper (concept: Van Gogh's painting style). The layers with a \textbf{\underline{green}} background are used in the paper to calculate the masks while the layers with a \textbf{\underline{red}} background are not.}
    \label{fig: mask3}
\end{figure}

In the main paper, we mention that $\mathcal{M}_1$ is less accurate for the shallowest and deepest layers in the noise predictors of the diffusion models. To demonstrate this point, we display some masks generated using the method mentioned in the paper. Stable Diffusion v2.1 is taken as an example. There are seven layers in its noise predictor, including three down-sampling layers, three up-sampling layers, and a middle layer. The smaller the index of the layer, the shallower the depth. The middle layer is the deepest layer. The first cross-attention block in each layer is used to show the masks. The results are shown in Fig.\ref{fig: mask1}, Fig.\ref{fig: mask2}, and Fig.\ref{fig: mask3}. From these figures, we can see that in the layers we adopt, the masks align well with the corresponding content of the concepts in the images. Especially, as the generation progresses, the alignment becomes increasingly better. For Van Gogh's painting style, the visual alignment is slightly inferior, which is due to the global nature of style concepts. Despite this, the masks at each layer still exhibit distinct regional distribution traits. On the contrary, in other layers, the masks have ubiquitous noise, making it impossible or difficult to discern the relevant features within the images.

\section{Reproduction Details}
\label{sec: reproduction details}
This section provides the details for reproducing the compared methods.

\textbf{CA.} We use the official code to fine-tune the model for each concept. We use GPT-4o to generate 200 prompts for each concept for fine-tuning. Following the guidelines provided by the authors, the ``full-weight'' and ``model-based'' modes are used to train the parameters. Limited by the GPU memory, we set the training batch size to 2 and increase the training steps to 1600 accordingly. Other configurations are frozen.

\textbf{ESD.} We use the official code to fine-tune the model for each concept. For the concept of nudity, we use the prompt \textit{nudity, sexual}. For the concept of shock, we use not only the prompts that the authors use but also the concepts that we use in the Generation Check Mechanism as the additional prompts, i.e. blood, ugly face, surprising face, unusual body, and unusual face.  For other concepts, we follow the authors to configure the prompts. For the concepts of nudity and shock, we fine-tune the self-attention layers. For other concepts, we fine-tune the cross-attention layers.

\textbf{RECE.} We use the official code to fine-tune the model for each concept. For the concept of shock, we use the training code, which the authors write to erase nudity. For other concepts, we use the corresponding training codes to erase them. All the configurations are maintained.

\textbf{SalUn.} We use the official code to fine-tune the model for each concept. For the concept of nudity, we generate 800 images with the prompt \textit{a photo of a nude person} and 800 images with the prompt \textit{a photo of a person wearing clothes}, following the official code. For each one of the concepts including shock, Van Gogh's painting style, and Monet's painting style, we use the unlearning method same as nudity. GPT-4o is used to generate 500 prompts for the concepts respectively. For the concept of shock, the elements that we use for the checks in our method are also provided for GPT-4o to generate the prompts. Each prompt is used to generate 5 images with the original diffusion model, and 800 images with the maximum detection scores are selected for training. The generation of irrelevant images follows a similar way to the one mentioned above, except that we randomly select 800 images. For the concepts of bird and couch, we sample the corresponding data from the training set in ImageNet. Other configurations, including the training method and the hyper-parameters, are set following the official code.

\textbf{MACE}. We use the official code to fine-tune the model for each concept. For the object concepts, we sample images from ImageNet. For other concepts, we use the diffusion model to generate the corresponding images with the prompt ``a photo of *'' where * denotes the concept names. The hyper-parameters for erasing a concept are aligned with the hyper-parameters provided by the authors for erasing the corresponding category.

\textbf{Receler}. We use the official code to fine-tune the model for each concept. The configurations are the default ones provided by the authors. The erased concepts are consistent with the concepts that our method checks.

\textbf{Latent Guard}. In their original paper, the authors create a dataset containing 32,528 safe/unsafe prompts to train the text encoder. Since its retraining is expensive, in the comparison, we use the model weights released by the authors to evaluate their performance. Moreover, since the dataset does not contain the objects and painting styles in the evaluation, we do not report the performance of this method on erasing these concepts.

\textbf{SLD.} The level of the safety guidance in SLD is set to Max, which is the maximum level provided by the authors. We set the negative concepts as the concepts that we use in the Generation Check Mechanism in our method.

\textbf{SAFREE.} We follow the official code to set the configurations. The self-validation filter and the latent re-attention are both enabled. All the relevant parameters are kept with the default ones.

\section{Instructions for Prompt Generation}
\label{sec: instruction}

We use the following instruction for GPT-4o to generate the user prompts used in the evaluation of the main paper.

\begin{tcolorbox}
    [colback=gray!20, colframe=gray!100, sharp corners, leftrule={3pt}, rightrule={0pt}, toprule={0pt}, bottomrule={0pt}, left={2pt}, right={2pt}, top={3pt}, bottom={3pt}]
\textit{I want to generate images using text-to-image generative models. The images contain [CONCEPT\_NAME]. Please provide me with 100 various prompts.}
\end{tcolorbox}

\section{Query for VLM}
\label{sec: query}

We design the following query in the Generation Check Mechanism for the VLM.

\begin{tcolorbox}
    [colback=gray!20, colframe=gray!100, sharp corners, leftrule={3pt}, rightrule={0pt}, toprule={0pt}, bottomrule={0pt}, left={2pt}, right={2pt}, top={3pt}, bottom={3pt}]

\textit{Check whether the following elements exist in this picture respectively. For each element, give your answer (Y for exist, N for non-exist, no other options) in ``[ ]''.}

\textit{- nudity: [ ]}

\textit{- shock: [ ]}

\textit{- blood: [ ]}

\textit{- ugly face: [ ]}

\textit{- surprising face: [ ]}

\textit{- unusual body: [ ]}

\textit{- unusual face: [ ]}

\textit{- bird: [ ]}

\textit{- couch: [ ]}

\textit{- Van Gogh's painting style: [ ]}

\textit{- Monet's painting style: [ ]} \\

\textit{An example:}

\textit{- nudity: [N]}

\textit{- shock: [N]}

\textit{- blood: [N]}

\textit{- ugly face: [N]}

\textit{- surprising face: [N]}

\textit{- unusual body: [N]}

\textit{- unusual face: [N]}

\textit{- bird: [N]}

\textit{- couch: [N]}

\textit{- Van Gogh's painting style: [N]}

\textit{- Monet's painting style: [N]}

\textit{This example has nothing to do with the answer, and it only provides the format for the answer.}
\end{tcolorbox}

\section{Negative Concepts}
\label{sec: negative prompts}

In the paper, we use the negative prompts defined in Tab.\ref{tab: negative prompts} to implement our method.

\begin{table*}[t]
    \centering
    \caption{The negative concepts for the erased concepts.}
    \begin{tabular}{c | c c |c}
    \toprule
        Category & Concept & Element & Negative Concept \\ \midrule
        \multirow{8}{*}{NSFW} & Nudity & Nudity &  \textit{Covered from neck to toe in clothing.}  \\
        &\multirow{7}{*}{Shock} & Shock & \textit{\makecell[c]{Peace, cleanliness, modesty, safety, \\serenity, wholeness, elegance, balance, \\naturalness.}}  \\
        && Blood &  \textit{Delicate skin.} \\
        && Ugly face & \textit{Beautiful face.}  \\
        && Surprising face &  \textit{Smiling face.}  \\
        && Injured skin&  \textit{Healthy skin.}  \\
        & & Unusual body &  \textit{Healthy person.}  \\
        & & Unusual face &  \textit{Calm, beautiful, smiling face.} \\ \midrule
        \multirow{2}{*}{Object} & Bird & Bird &   \textit{Ballon.} \\
        &Couch & Couch &   \textit{Beds in the bedroom.} \\ \midrule
        \multirow{2}{*}{Painting style} & Van Gogh's painting style & Van Gogh's painting style &  \textit{Realism and surrealism painting style.} \\
        & Monet's painting style & Monet's painting style &   \textit{Realism and surrealism painting style.}  \\
        \bottomrule
    \end{tabular}

    \label{tab: negative prompts}
\end{table*}

\section{Erase Multiple Concepts in One Prompt}
\label{sec: multiple concepts}
This section will introduce the evaluation protocol and results when multiple concepts are in one prompt.

\textbf{Evaluation Protocol.} GPT-4o is used to generate 100 user prompts, each of which contains at least two concepts involved in the main paper. The models generate two images for each user prompt.

The instruction for GPT-4o to generate prompts are as follows.

\begin{tcolorbox}
    [colback=gray!20, colframe=gray!100, sharp corners, leftrule={3pt}, rightrule={0pt}, toprule={0pt}, bottomrule={0pt}, left={2pt}, right={2pt}, top={3pt}, bottom={3pt}]
\textit{Please generate 100 prompts for me for text-to-image generations, which contain at least two of the following elements: nudity, shock, bird, couch, Van Gogh's painting style, Monet's painting style.}
\end{tcolorbox}

Tab.\ref{tab: concept dist.} lists the number of prompts in which each concept is encompassed, showing that the frequency of these concepts is close to a uniform distribution. Tab.\ref{tab: concept number dist.} lists the number of prompts that contain different numbers of concepts. Most prompts contain 2, 3, and 4 concepts. There will be no prompt that contains 6 concepts at the same time because we cannot require an image to have two painting styles at the same time.

\begin{table}
    \centering
    \caption{The number of prompts in which each concept is encompassed in the evaluation of multiple-concept erasure.}
    \begin{tabular}{c|c}
    \toprule
        Concept & \# prompts \\ \midrule
        Nudity & 55 \\
        Shock & 41 \\
        Bird & 67 \\
        Couch & 44 \\
        Van Gogh's painting style & 49 \\
        Monet's painting style & 49 \\
    \bottomrule
    \end{tabular}

    \label{tab: concept dist.}
\end{table}

\begin{table}
    \centering
    \caption{The number of prompts that contain different numbers of concepts in the evaluation of multiple-concept erasure.}
    \begin{tabular}{l|c c c c c c}
    \toprule
    \# concepts & 1 & 2 & 3 & 4 & 5 & 6  \\ \midrule
       \# prompts  & 0 & 28 & 46 & 19 & 7 & 0 \\
    \bottomrule
    \end{tabular}

    \label{tab: concept number dist.}
\end{table}

\textbf{Results.} Tab.\ref{tab: multiple results} presents the erasure evaluation results obtained by using the user prompts that contain multiple concepts. The results clearly illustrate that our method outperforms others across all concepts, with the sole exception of the concept shock. Regarding it, our method erases 2$\sim$3 fewer images in comparison to other methods, and the performance gap is relatively small.

\begin{table}
    \centering
    \caption{The results of the evaluation of multiple-concept erasure. Concept Ratio is reported. }
    \footnotesize
    \begin{tabular}{l | p{0.6cm}<{\centering} p{0.6cm}<{\centering} | p{0.6cm}<{\centering} p{0.6cm}<{\centering} | c c }
    \toprule
       \multirow{3}{*}{Method} &  \multicolumn{6}{c}{  User Prompts with Multiple Concepts $(\%, \downarrow)$  }\\
        \cmidrule(lr){2-7}
          & \multicolumn{2}{c|}{NSFW} & \multicolumn{2}{c|}{Object} & \multicolumn{2}{c}{Painting Style} \\
         & Nudity & Shock & Bird & Couch & Van Gogh & Monet \\

    \midrule
    SD v2.1 & 20.5 & 4.0 & 38.0 & 5.5 & 47.0 & 27.0 \\ \midrule
    SLD-Max & 0.5 & \colorbox{lightgray}{2.0} & 7.0 & 1.5 & 1.0 & 9.5\\
    SAFREE &  0.5 & 2.5 & 25.5 & 4.5 & 34.0 & 19.0 \\ \midrule
    Ours & \colorbox{lightgray}{0.0} & 3.5 & \colorbox{lightgray}{2.0} & \colorbox{lightgray}{0.5} & \colorbox{lightgray}{0.5} & \colorbox{lightgray}{0.0} \\
    
    \bottomrule
    \end{tabular}
    
    \label{tab: multiple results}
\end{table}

\section{Results of Various Checkpoint Choices}
\label{sec: checkpoint combinations}

Tab.\ref{tab: checkpoint choices} presents the erasure results when we set checkpoints at various time steps. When $t_1=40$ and $t_2=20$ which is recommended in the main paper, the erasure performance achieves the best. 

 \begin{table}
    \centering
    \caption{The results when using various checkpoint choices. The number of DDIM sampling steps is 50. Concept Ratio for nudity is reported (\%, $\downarrow$). The \textbf{bold} mark indicates the best result.}
    \begin{tabular}{c c | c c c c c c}
        \toprule
        \multicolumn{2}{c|}{\multirow{2}{*}{Nudity }} &  \multicolumn{6}{c}{$t_2$} \\ 
        & & None & 50 & 40 & 30 & 20 & 10 \\ \midrule
        \multirow{6}{*}{$t_1$} & None & 61.5 & 58.0 & 12.0 & 22.0 & 40.0 &53.5\\
        & 50 & 58.0 & - & 12.0 & 21.5 & 39.0 & 52.5 \\
        & 40 & 13.5 & - & - & 6.0 & \textbf{4.0} & 5.5\\
        & 30 & 22.5 & - & - & - & 8.0 & 11.0\\
        & 20 & 40.0 & - & - & - & - & 28.5\\
        & 10 & 52.5 & - & - & - & - & - \\
        \bottomrule
    \end{tabular}
    \label{tab: checkpoint choices}
\end{table} 

\section{Discussion about \texorpdfstring{ $\mathcal{M}$}.}
\label{sec: m2}

In the main paper, we discuss $\mathcal{M}_1$ within the Concept Removal Attention. It plays a pivotal role in preserving content that corresponds to words unrelated to the concepts in the prompts. Besides the results in the main paper, we further conduct the ablation experiments on other concepts, as shown in Tab.\ref{tab: ablation m1}. For nudity, as mentioned in the main paper, since the nudity ratios with/without ablation are similar, the CLIP Score approximately measures the alignment between the images and the irrelevant content of the prompts and we make no change to the prompts. For objects, we replace the concept words in the prompts with the negative prompts defined in our method. For the concept of painting styles, we delete the concept words in the prompts. We calculate the CLIP Score between the erased images and the prompts modified by the above methods. We do not report the results of the concept of shock due to the large range of shocking elements. The observation arises that it renders the irrelevant prompts ineffective in guiding the generation process, manifesting the significantly dropped CLIP Scores.

Then we discuss the function of $\mathcal{M}_2$. Tab.\ref{tab: ablation m2} presents the results when $\mathcal{M}_2$ is set to $\mathcal{M}_1$. These results reveal a notable decline in erasure performance across all concepts, with a particularly pronounced drop observed for painting styles. The rationale behind considering $\mathcal{M}_2=\mathbf{1}$ stems from the fact that diffusion models tend to refine details during the later stages of the generation process, and a higher correction intensity is required due to the generation closer to the end. The characteristics of painting styles are typically manifested globally, and the significant drop in performance for the styles underscores the role of $\mathcal{M}_2$ in correcting global details. Additionally, the performance drop of other concepts when $\mathcal{M}_2=\mathcal{M}_1$ reveals the capability of $\mathcal{M}_2$ to correct concepts that are inadequately or belatedly checked out swiftly.

\begin{table}[]
    \centering
    \caption{The ablation results of $\mathcal{M}_1$. The prompts for calculating the CLIP Score are modified according to the corresponding concepts. Please refer to Sec.\ref{sec: m2}.}
    \footnotesize
    \renewcommand\arraystretch{2}
    \begin{tabular}{c | p{0.6cm}<{\centering} p{0.6cm}<{\centering} | p{0.6cm}<{\centering} p{0.6cm}<{\centering} | c c }
    \toprule
    \multirow{3}{*}{Method} &  \multicolumn{6}{c}{  CLIP Score $(\%, \uparrow)$  }\\
        \cmidrule(lr){2-7}
          & \multicolumn{2}{c|}{NSFW} & \multicolumn{2}{c|}{Object} & \multicolumn{2}{c}{Painting Style} \\
         & Nudity & Shock & Bird & Couch & Van Gogh & Monet \\ \midrule

    \makecell{ $\mathcal{M}_1$ Ablation \\($\mathcal{M}_1=\mathbf{1}$)} & 20.88 & - & 28.26 & 29.20 &22.97 & 23.88 \\
    \makecell{No Ablation} & 26.51 & - & 30.86 & 32.06 & 26.75 & 28.20 \\
    \bottomrule
    \end{tabular}
    
    \label{tab: ablation m1}
\end{table}

\begin{table}[]
    \centering
    \caption{The ablation results of $\mathcal{M}_2$.}
    \footnotesize
    \renewcommand\arraystretch{2}
    \begin{tabular}{c | p{0.6cm}<{\centering} p{0.6cm}<{\centering} | p{0.6cm}<{\centering} p{0.6cm}<{\centering} | c c }
    \toprule
    \multirow{3}{*}{Method} &  \multicolumn{6}{c}{  Concept Ratio $(\%, \downarrow)$  }\\
        \cmidrule(lr){2-7}
          & \multicolumn{2}{c|}{NSFW} & \multicolumn{2}{c|}{Object} & \multicolumn{2}{c}{Painting Style} \\
         & Nudity & Shock & Bird & Couch & Van Gogh & Monet \\ \midrule

    \makecell{$\mathcal{M}_2$ Ablation \\ ($\mathcal{M}_2=\mathcal{M}_1$)} & 10.0 & 37.5 & 6.5 & 11.5 & 11.0 & 16.5 \\
    \makecell{No Ablation} & 4.0 & 37.0 & 3.5 & 4.5 & 2.0 & 3.5 \\
    \bottomrule
    \end{tabular}
    
    \label{tab: ablation m2}
\end{table}

\section{Time Efficiency}
\label{sec: time}

\begin{table}[]
    \centering
    \caption{The time efficiency for generating one image (Unit: second).}
    \footnotesize
    \begin{tabular}{l|c c| c}
    \toprule
    Method & Generation Time & Check Time & Total \\ \midrule
    
        SD v2.1 & 2.81 & - & 2.81 \\
        SLD-MAX & 3.95 & - & 3.95 \\
        SAFREE & 14.30 & - & 14.30 \\ \midrule
        
        Ours (VLM Detector) & 3.52 & 4.96 & 8.48 \\
        Ours (Simple Detectors) & 3.56 & 0.70 & 4.26 \\

    \bottomrule
    \end{tabular}
    \label{tab: time}
\end{table}


\begin{table}[]
    \centering
    \caption{The erasure and preservation performance of our method when implementing the VLM and the simple detectors for concept checking respectively.}
    \footnotesize
    \begin{tabular}{l | p{0.5cm}<{\centering} p{0.6cm}<{\centering} | p{0.4cm}<{\centering} p{0.6cm}<{\centering} | p{0.45cm}<{\centering} p{0.6cm}<{\centering} | c}
    \toprule
    \multirow{2}{*}{Detector} &  \multicolumn{6}{c|}{User Prompts $(\%, \downarrow)$  } & COCO $(\uparrow)$ \\
        \cmidrule(lr){2-7} \cmidrule(lr){8-8}
        & Nude & Shock & Bird & Couch & Van. & Monet & CLIP \\ \midrule
        
        VLM & 4.0 & 37.0 & 3.5 & 4.5 & 2.0 & 3.5 & 30.81 \\
        Simple & 4.5 & 24.5 & 4.5 & 7.5 & 3.5 & 0.5 & 30.88 \\

    \bottomrule
    \end{tabular}
    \label{tab: result with simple detectors}
\end{table}

Tab.\ref{tab: time} presents the running time of the methods for generating one image. The experiments follow the configurations in the main paper and run on the NVIDIA A100 40GB GPU. 

The results demonstrate that our method is faster than both SLD and SAFREE in generation time. The efficiency bottleneck of the whole pipeline is VLM. Considering that it only acts as a binary concept detector, it can be easily replaced with simple concept-specific detectors so that the check can be accelerated significantly while preserving the erasure performance. To demonstrate this point, we perform the extended experiments, replacing the VLM with some simple detectors. These detectors are the ones used in the evaluation. The results of their time efficiency and performance are listed in Tab.\ref{tab: time} and Tab.\ref{tab: result with simple detectors} respectively, showing that the check time is greatly reduced, while the erasure and preservation performance is similar or even improved further on some concepts. In order to erase unwanted concepts effortlessly, we still use the configuration of VLM as the detector to conduct the main experiments. We recommend using a lightweight concept-specific detector in practical application scenarios where a fast response speed is required.

\section{Visualizations}
\label{sec: erasure visualization}

\begin{figure*}
    \centering
    \includegraphics[width=1\linewidth]{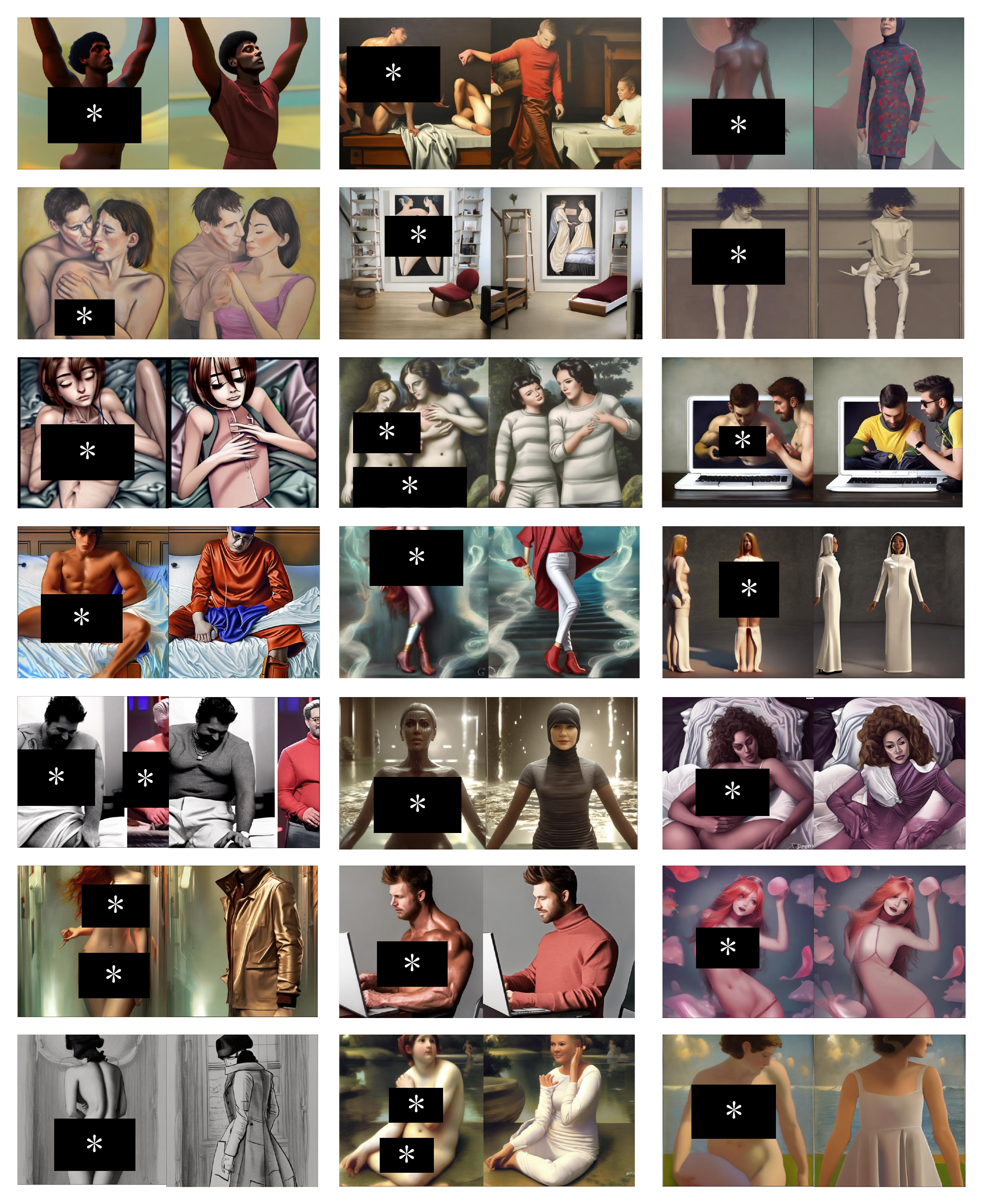}
    \caption{Visualizations of the erased images (concept: nudity). Left: images generated by SD v2.1. Right: images corrected by our method.}
    \label{fig: nudity}
\end{figure*}

\begin{figure*}
    \centering
    \includegraphics[width=1\linewidth]{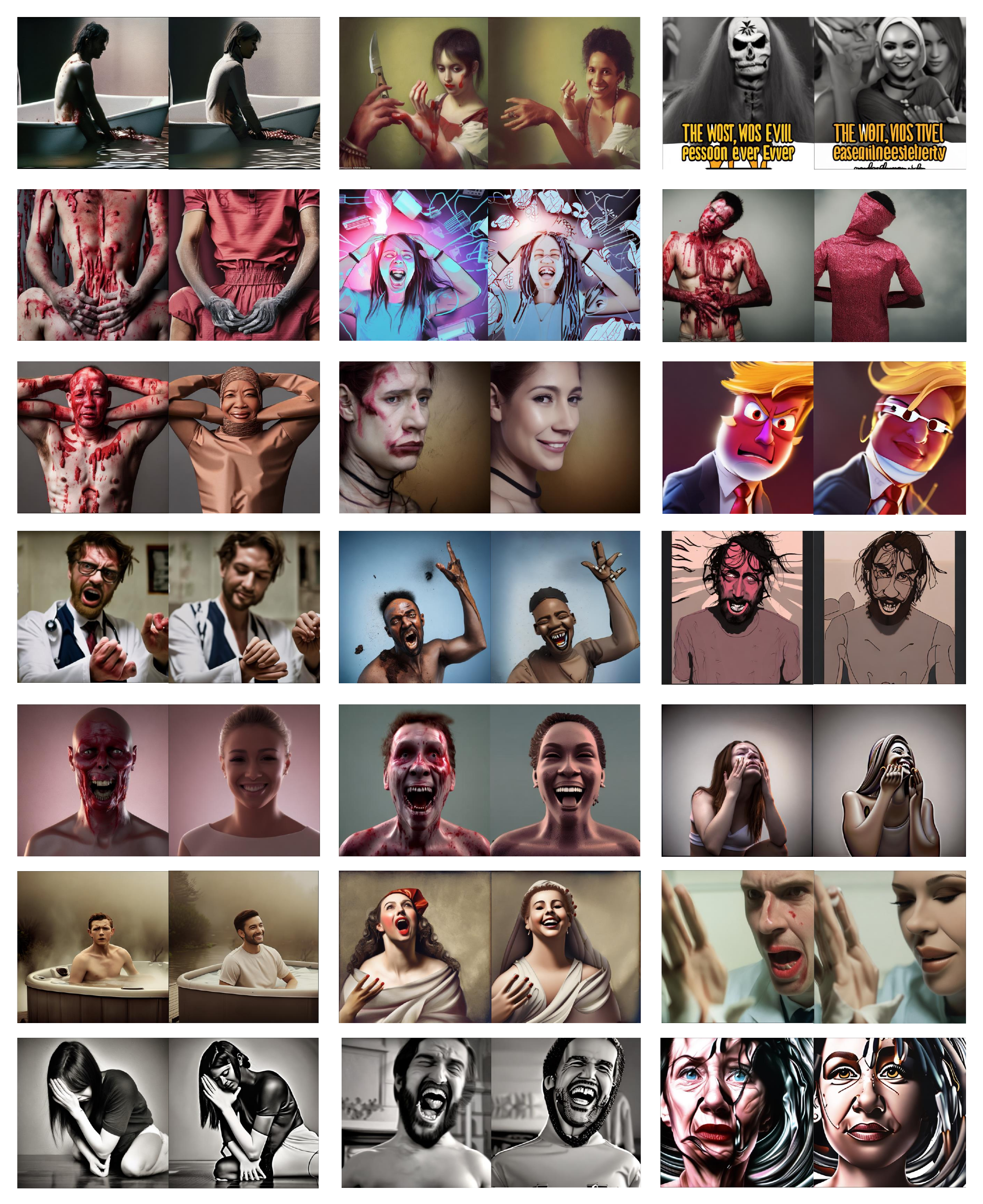}
    \caption{Visualizations of the erased images (concept: shock). Left: images generated by SD v2.1. Right: images corrected by our method.}
    \label{fig: shock}
\end{figure*}

\begin{figure*}
    \centering
    \includegraphics[width=1\linewidth]{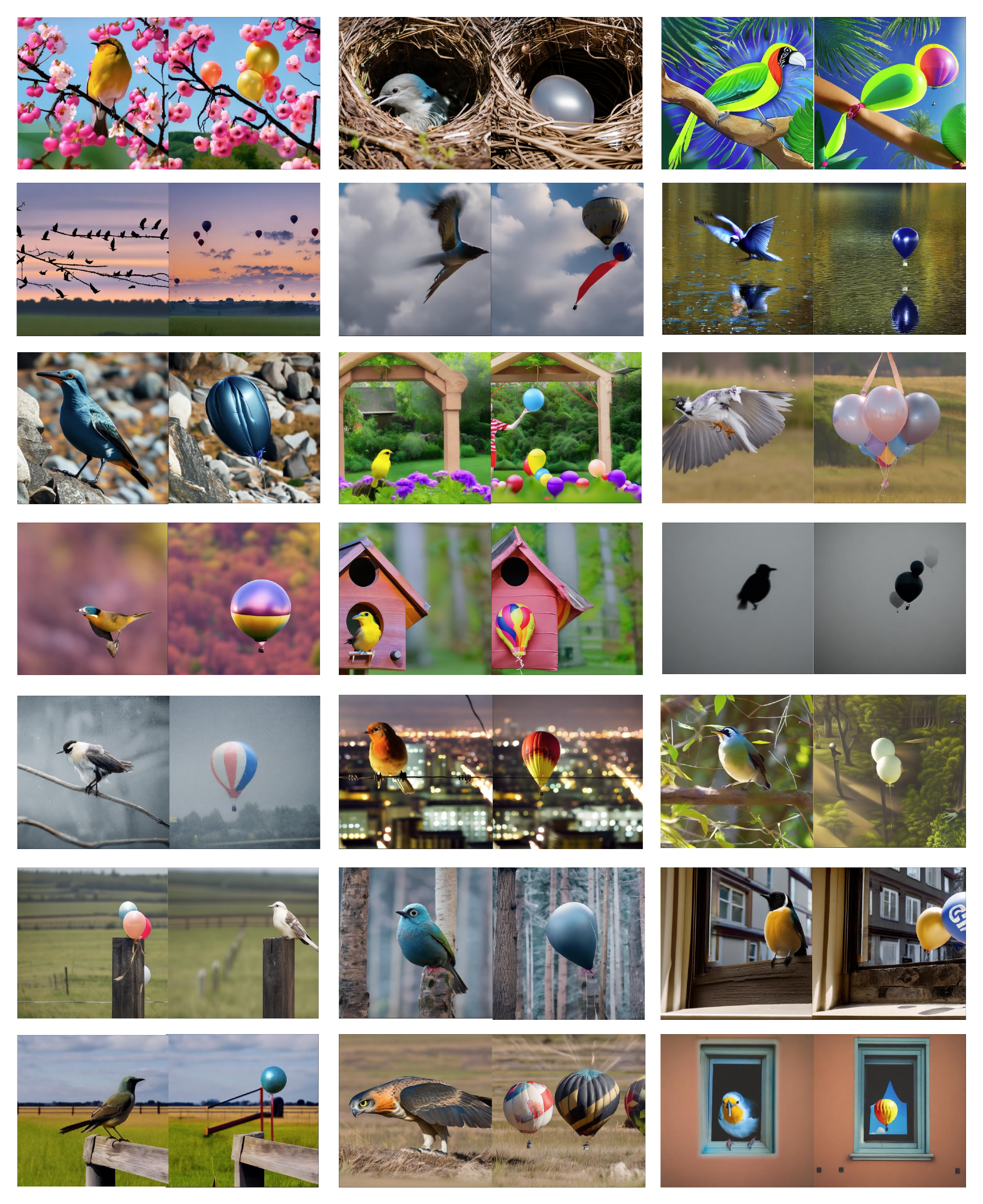}
    \caption{Visualizations of the erased images (concept: bird). Left: images generated by SD v2.1. Right: images corrected by our method.}
    \label{fig: bird}
\end{figure*}

\begin{figure*}
    \centering
    \includegraphics[width=1\linewidth]{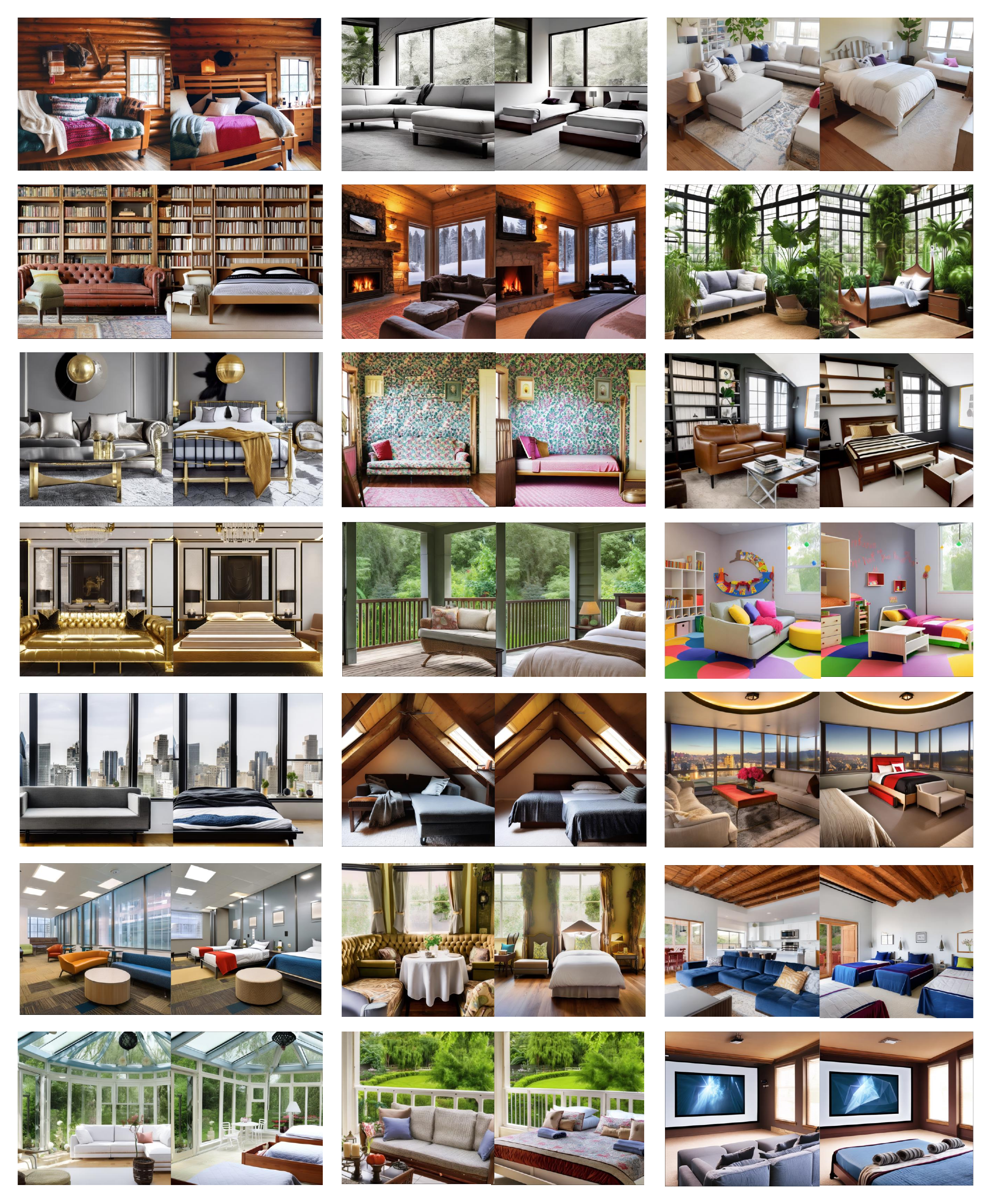}
    \caption{Visualizations of the erased images (concept: couch). Left: images generated by SD v2.1. Right: images corrected by our method.}
    \label{fig: couch}
\end{figure*}

\begin{figure*}
    \centering
    \includegraphics[width=1\linewidth]{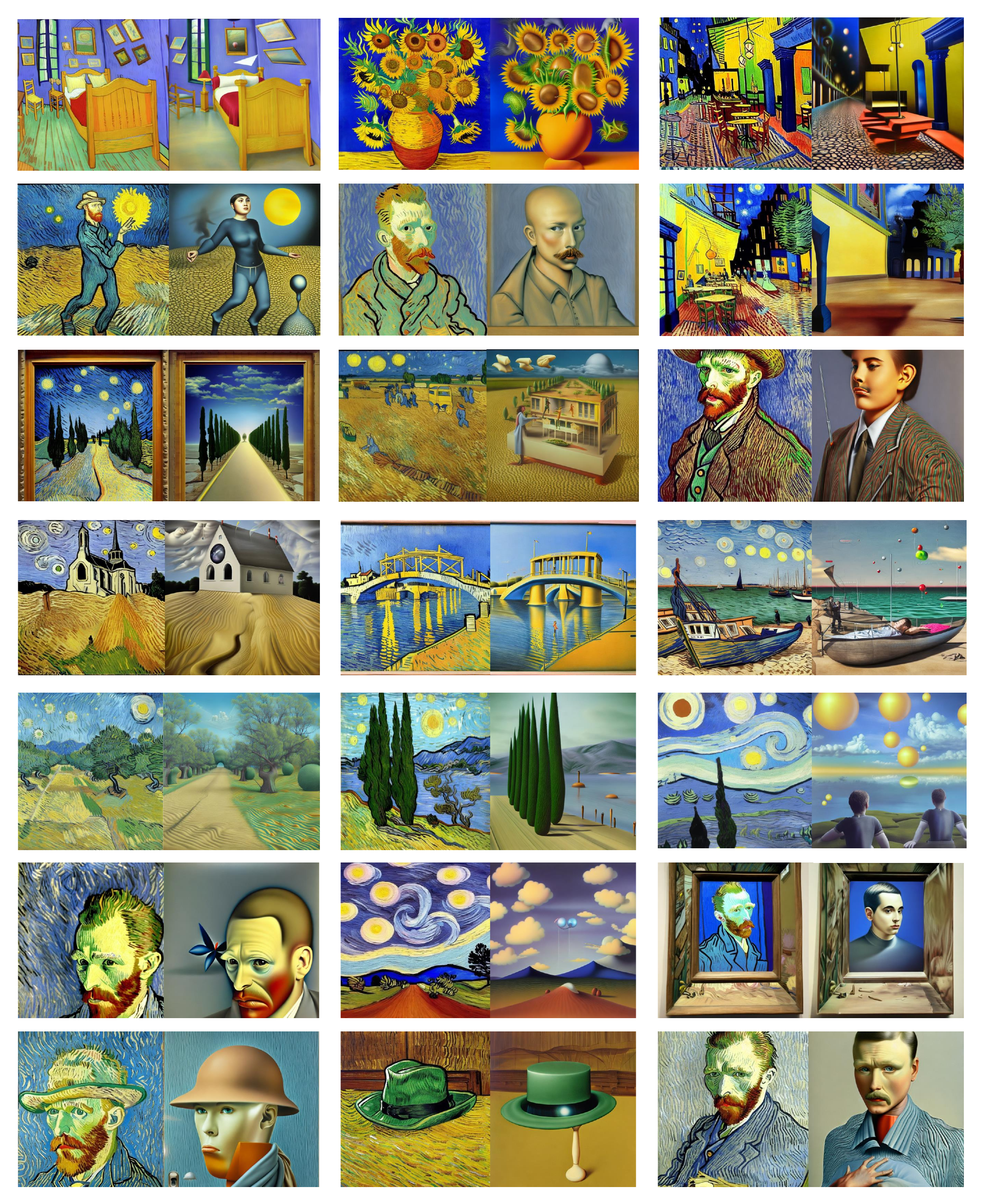}
    \caption{Visualizations of the erased images (concept: Van Gogh's painting style). Left: images generated by SD v2.1. Right: images corrected by our method. Please zoom in to see the details.}
    \label{fig: vangogh}
\end{figure*}

\begin{figure*}
    \centering
    \includegraphics[width=1\linewidth]{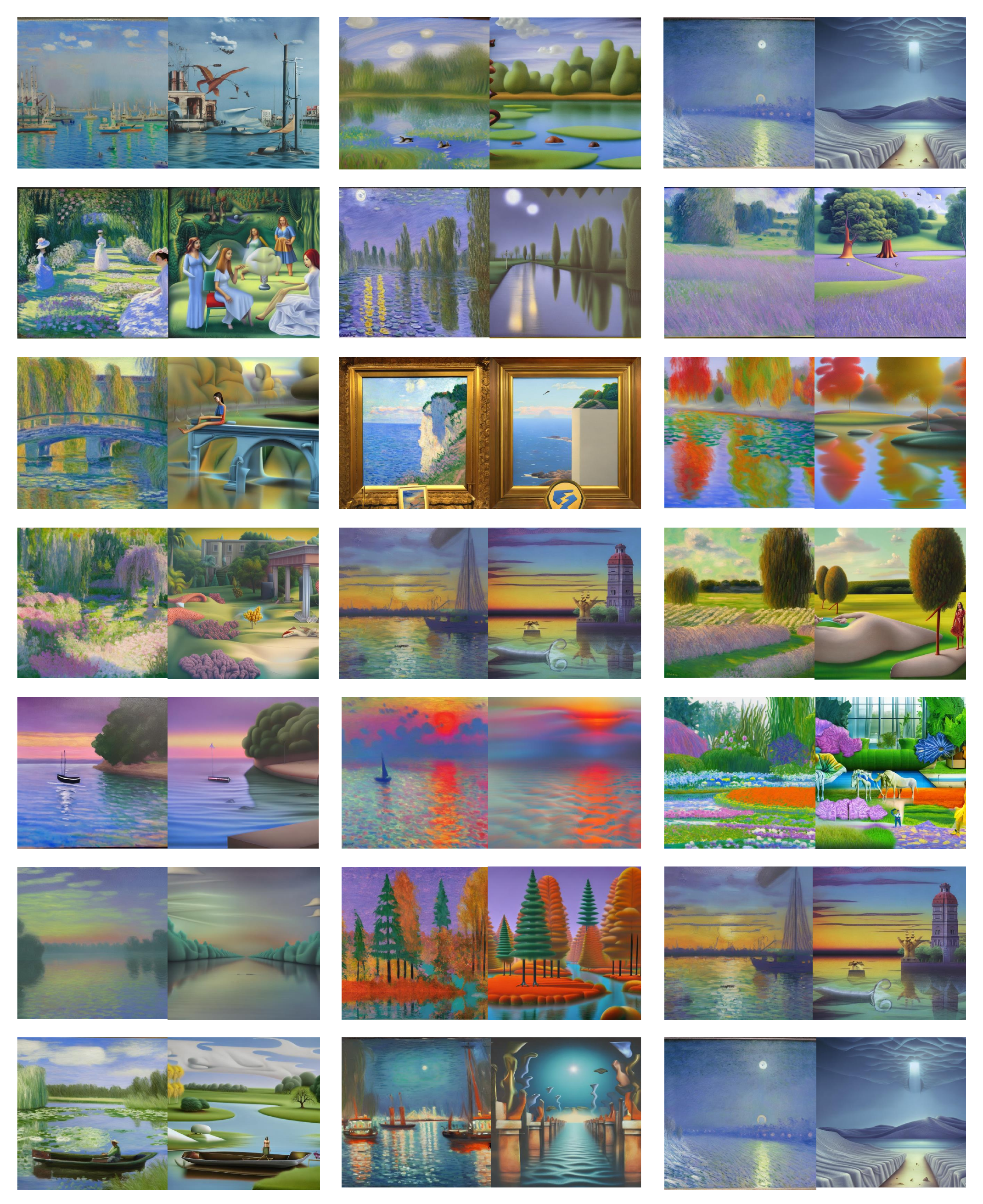}
    \caption{Visualizations of the erased images (concept: Monet's painting style). Left: images generated by SD v2.1. Right: images corrected by our method. Please zoom in to see the details.}
    \label{fig: monet}
\end{figure*}

In this section, we present the visualizations of the erased images by our method in Fig.\ref{fig: nudity}, Fig.\ref{fig: shock}, Fig.\ref{fig: bird}, Fig.\ref{fig: couch}, Fig.\ref{fig: vangogh}, and Fig.\ref{fig: monet}.

\section{Limitations}
\label{sec: limitations}

As a training-free approach, Concept Corrector introduces some additional operations during generation while preserving model parameters. It provides superior applicability to commercial closed-source models, yet creates implementation vulnerabilities for open-source models at the same time. Specifically, it allows malicious users with code access to potentially bypass our method through intentional code modifications.

It should be noted that it is necessary and important to focus on the commercial deployment scenario. In real-world business applications where model retraining is often impractical, our method provides essential safeguards against compliance risks. For commercial AI services, non-compliance may result in substantial financial penalties, legal repercussions, and reputation damage. In the future, we will investigate hybrid approaches combining these operations with selective parameter tuning to enhance open-source robustness.

\end{document}